\documentclass[sigconf]{acmart} 
\usepackage{booktabs}
\usepackage[table]{xcolor}
\usepackage{colortbl}
\usepackage{caption}
\usepackage{makecell}
\usepackage{graphicx}
\usepackage{subcaption}
\usepackage{booktabs}
\usepackage{graphicx}
\usepackage{multirow}
\usepackage{float}
\usepackage{caption}
\usepackage[most]{tcolorbox}
\usepackage{float}
\usepackage{booktabs}
\usepackage{tabularx}
\usepackage{booktabs}
\usepackage{array}
\usepackage{caption}
\usepackage{needspace}
\usepackage[most]{tcolorbox}
\usepackage{placeins}
\usepackage{graphicx}
\usepackage{tikz}
\usepackage{url}

\usepackage{mathtools}   
\usepackage{microtype}   
\usepackage{cuted}   
\usetikzlibrary{positioning,arrows.meta,calc}

\usepackage[ruled,vlined]{algorithm2e}
\usepackage{xcolor}
\usepackage{colortbl} 

\definecolor{easytxt}{RGB}{200, 255, 200}   
\definecolor{medtxt}{RGB}{255, 255, 200}    
\definecolor{hardtxt}{RGB}{255, 200, 200}   

\newcommand{\easycell}{\cellcolor{easytxt}}
\newcommand{\medcell}{\cellcolor{medtxt}}
\newcommand{\hardcell}{\cellcolor{hardtxt}}
\tcbuselibrary{breakable}

\AtBeginDocument{%
  }

\setcopyright{acmlicensed}
\copyrightyear{2018}
\acmYear{2018}
\acmDOI{XXXXXXX.XXXXXXX}
\acmConference[Conference acronym 'XX]{Make sure to enter the correct
  conference title from your rights confirmation email}{June 03--05,
  2018}{Woodstock, NY}
\acmISBN{978-1-4503-XXXX-X/2018/06}

\begin{document}

\title{%
  \raisebox{-0.35\height}{\includegraphics[height=2.05em]{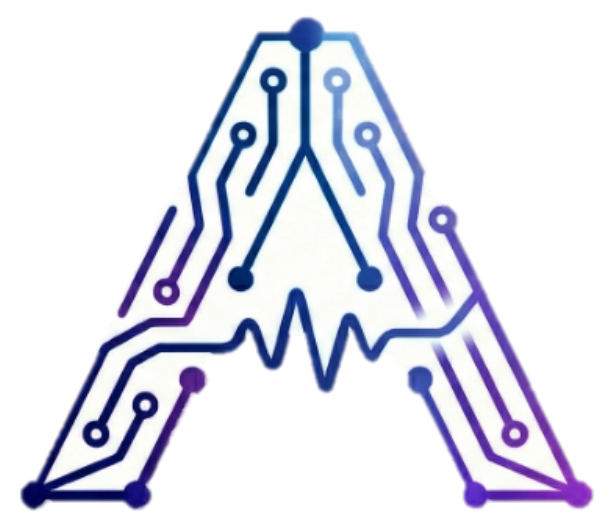}}%
  AnalogAgent: Self-Improving Analog Circuit Design Automation with LLM Agents
}

\author{Zhixuan Bao}
\affiliation{%
  \institution{Nanyang Technological University}
  \city{}
  \country{Singapore}
}\authornote{Work done during internship at A*STAR.}
\email{zhixuan005@e.ntu.edu.sg }

\author{Zhuoyi Lin}
\affiliation{%
  \institution{Institute for Infocomm Research}
  \city{}
  \country{Singapore}
}\authornote{Corresponding author.}
\email{Lin_Zhuoyi@a-star.edu.sg}

\author{Jiageng Wang}
\affiliation{%
 \institution{Nanyang Technological University}
 \country{Singapore}
 }
\email{WANG2396@e.ntu.edu.sg}

\author{Jinhai Hu}
\affiliation{%
  \institution{Institute of Microelectronics}
  \country{Singapore}}
\email{hujh@a-star.edu.sg}

\author{Yuan Gao}
\affiliation{%
  \institution{Institute of Microelectronics}
  \country{Singapore}}
\email{gaoy@a-star.edu.sg}

\author{Yaoxin Wu}
\affiliation{%
  \institution{Eindhoven University of Technology}
  \city{Eindhoven}
  \country{Netherlands}
}
\email{y.wu2@tue.nl}

\author{Xiaoli Li}
\affiliation{%
  \institution{Singapore University of Technology}
  \country{and Design, Singapore}}
\email{xiaoli_li@sutd.edu.sg}

\author{Xun Xu}
\affiliation{%
  \institution{Institute for Infocomm Research}
  \country{Singapore}}
\email{Xu_Xun@a-star.edu.sg}

\renewcommand{\shortauthors}{Bao et al.}

\begin{abstract}

Recent advances in large language models (LLMs) suggest strong potential for automating analog circuit design. Yet most LLM-based approaches rely on a single-model loop of generation, diagnosis, and correction, which favors succinct summaries over domain-specific insight and suffers from context attrition that erases critical technical details.
To address these limitations, we propose \textbf{AnalogAgent}, a training-free agentic framework that integrates an LLM-based multi-agent system (MAS) with self-evolving memory (SEM) for analog circuit design automation. AnalogAgent coordinates a Code Generator, Design Optimizer, and Knowledge Curator to distill execution feedback into an adaptive playbook in SEM and retrieve targeted guidance for subsequent generation, enabling cross-task transfer without additional expert feedback, databases, or libraries. Across established benchmarks, AnalogAgent achieves 92\% Pass@1 with Gemini and 97.4\% Pass@1 with GPT-5. Moreover, with compact models (e.g., Qwen-8B), it yields a +48.8\% average Pass@1 gain across tasks and reaches 72.1\% Pass@1 overall, indicating that AnalogAgent substantially strengthens open-weight models for high-quality analog circuit design automation.

\end{abstract}


\received{20 February 2007}
\received[revised]{12 March 2009}
\received[accepted]{5 June 2009}

\maketitle
\captionsetup{labelfont=normalfont,textfont=normalfont}

\section{Introduction}

Analog circuit design is a cornerstone of modern scientific discovery, enabling the reliable translation of physical phenomena into calibrated electrical signals that can be measured, controlled, and computed with high fidelity. It underpins core instrumentation across biomedical sensing, environmental monitoring, and large-scale experimental facilities, where noise, accuracy, and stability directly shape data quality and reproducibility \cite{gray2024analysis}.
In contrast to digital design, which benefits from standardized abstractions and mature automation pipelines, analog design relies heavily on expert knowledge, heuristic reasoning, and iterative simulation-driven refinement. Although conventional electronic design automation (EDA) tools have achieved progress in areas such as device sizing and topology optimization \cite{wang2020gcn,budak2023apostle}, they are typically limited to predefined architectures \cite{poddar2024data}, require extensive manual configuration \cite{liu2009analog}, and scale poorly when exploring novel circuit structures or heterogeneous design objectives \cite{mcconaghy2007simultaneous,palmers2009massively}. 

\begin{figure}[t]
  \centering
  \includegraphics[width=0.98\linewidth]{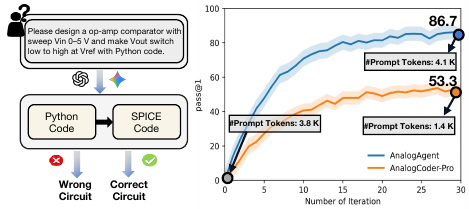}
  \caption{\textbf{Illustration of context attrition on the hard Task 25.} As context accumulates across iterations, LLM agents tend to compress it into shorter, less informative summaries, diminishing instruction salience and degrading detailed guidance. AnalogCoder Pro takes $\sim$7 minutes to produce its first correct circuit, whereas AnalogAgent succeeds in 17.13 seconds with learned knowledge.}
  \label{fig:comparision}
  \vspace{-0.15in}
\end{figure}

Recent advancements in large language models (LLMs) have revitalized academic interest in analog circuit design automation. By leveraging their capabilities in code generation, structured reasoning, and broad generalization, LLMs offer an alternative to traditional workflows that rely primarily on hand-crafted rules or tightly coupled numerical optimization \cite{lai2025analogcoder,lai2025analogcoder_pro,vungarala2024spicepilot, yin2024ado}. 
Early evidence is provided by AnalogCoder, which shows the feasibility of generating executable design scripts for analog circuits in a code-centric pipeline \cite{lai2025analogcoder}.
AnalogCoder-Pro further introduces a multimodal feedback-enhanced design flow \cite{lai2025analogcoder_pro}. These efforts establish an important benchmark for LLM-driven analog design automation with training-free code generation.
However, LLM-based analog circuit design automation remains constrained by several challenges. \textbf{(1) Single LLM paradigm.} Most approaches rely on a single LLM to jointly handle code generation, error diagnosis, and iterative refinement, coupling heterogeneous subtasks within one monolithic loop.
\textbf{(2) Context attrition.} 
As shown in Figure \ref{fig:comparision}, LLM agents tend to frequently rewrite the detailed specification and prior findings during iterative refinement, which progressively erode salient technical details and degrade the performance.
\textbf{(3) Dependence on circuit libraries.} Most of the existing methods assume access to additional circuit databases (e.g., topology library, circuit textbooks, and literature) to scaffold circuit design, which limits applicability in settings where such resources are unavailable or proprietary.

In this paper, we introduce AnalogAgent, a training-free agentic framework for analog circuit design automation. Unlike prior LLM-based methods that center on a single agent augmented with circuit libraries, AnalogAgent integrates an LLM-based \textbf{multi-agent system (MAS)} with \textbf{self-evolving memory (SEM)} to adaptively execute design, evaluation, and knowledge reuse. 
Specifically, AnalogAgent coordinates three agents. (1) The Code Generator generates task-specific SPICE/Python design scripts. (2) The Design Optimizer evaluates candidates through simulation-driven checks and converts execution feedback into targeted corrections and reusable best practices.
(3) The Knowledge Curator distills optimization feedback into structured, reusable design knowledge and incrementally updates the \textbf{Adaptive Design Playbook} within SEM. The playbook then supplies targeted guidance to the Code Generator, grounding subsequent generations in accumulated experience. 
By coupling workflow decomposition with iterative execution feedback and structured memory updates, AnalogAgent converts outcomes into actionable guidance, reduces redundant experimentation, and accelerates convergence to specification-compliant designs.

In summary, the main contributions of this work are:
(1) \textbf{Conceptual}: AnalogAgent represents an early attempt at a training-free agentic framework for analog circuit design automation that operates without additional circuit libraries.
(2) \textbf{Algorithmic:} AnalogAgent employs a coordinated multi-agent system with Code Generation, Design Optimization, and Knowledge Curation agents to support structured reasoning and sustained self-improvement. Its self-evolving memory mechanism retrieves guidance from an Adaptive Design Playbook and distills validated heuristics into reusable rules, mitigating context attrition and improving robustness and cross-task generalization.
(3) \textbf{Empirical:} We evaluate AnalogAgent and strong baselines on analog circuit design tasks spanning multiple complexity levels. Results show consistent gains and successful adaptation to compact open-weight models (e.g., Qwen-8B).

\section{Related Work}

\subsection{Circuit Design Automation}
Conventional analog automation tools provide practical benefits, yet key constraints remain. Recent device-sizing methods improve sample efficiency \cite{wang2020gcn,settaluri2020autockt}, with further gains from learned optimization and parallelization \cite{budak2021dnn,budak2023apostle}; related learning-based sizing approaches report similar efficiency improvements \cite{choi2023reinforcement,li2021circuit}, and newer systems advance sample-efficient optimization \cite{oh2024cronus,poddar2025insight}. Despite this progress, these methods are largely confined to predetermined circuit architectures, consistent with earlier pipelines \cite{liu2009analog}, which limits transfer when meeting specifications requires architectural adaptation, alternative biasing, or changes in feedback structure; they work best when the topology class is known and the objective reduces to parameter selection within a fixed template. In contrast, topology exploration often incurs substantial simulation overhead \cite{mcconaghy2007simultaneous, palmers2009massively}, depends on expert-derived analytical formulations \cite{veselinovic1995flexible,mcconaghy2008automated,zhao2020automated}, and offers limited structural diversity \cite{maulik2002integer}, while risking invalid configurations \cite{lu2023high}. Costs are dominated by feasibility screening across many candidates, with frequent failures from bias infeasibility, missing DC paths, or non-convergent operating points; expert dependence further reduces portability across circuit families, regimes, or process assumptions. Moreover, automated synthesis approaches such as \cite{poddar2024data} often require substantial up-front setup, restricting deployment to predefined component or topology libraries, including block curation, interface/range specification, and constraint encoding for correctness and simulator stability, which raises the barrier to extending the system to new components, macro-blocks, or architectural motifs.

\subsection{LLM-based Circuit Design Automation}

Early LLM-based analog circuit design approaches largely rely on prompt-driven generation, using curated instructions and in-context exemplars to steer LLMs toward synthesizing SPICE or Python netlists, as demonstrated by SPICEPilot \cite{vungarala2024spicepilot}, Artisan \cite{chen2024artisan}, and Ado-llm \cite{yin2024ado}.
A complementary paradigm explores training-based solusions, improving circuit synthesis via supervised fine-tuning, reinforcement learning, or dataset augmentation. Representative methods include AnalogGenie \cite{gao2025analoggenie}, LaMAGIC \cite{chang2024lamagic}, and CKTGNN \cite{dong2023cktgnn}. While effective, these approaches typically require additional labeled data and may demand repeated retraining when generalizing to new circuit families.
To further enhance reliability and reduce iterative search cost, recent work introduces external knowledge resources such as topology libraries, circuit tool repositories, or structured databases to condition generation, including AnalogXpert \cite{zhang2025analogxpert}, AMSnet-KG \cite{shi2025amsnet}, LADAC \cite{liu2024ladac}, LayoutCopilot \cite{liu2025layoutcopilot}, AnalogCoder \cite{lai2025analogcoder}, and AnalogCoder-pro \cite{lai2025analogcoder_pro}.
In contrast, AnalogAgent is training-free and does not rely on an additional database or a pre-built circuit library. 
Instead, it leverages execution signals from simulation and checker diagnostics, distilling both failures and successes into reusable rules that are stored in SEM and retrieved to guide subsequent generations, thereby forming a cross-task self-improvement loop. 
Moreover, AnalogAgent maintains strong performance with smaller models that are suitable for local or self-hosted deployment, reducing reliance on large proprietary backbones while retaining reliable first-pass synthesis and iterative refinement for industry companies.
\captionsetup{labelfont=normalfont,textfont=normalfont}
\begin{figure*}[!ht]
  \centering
  \includegraphics[width=1.0\linewidth]{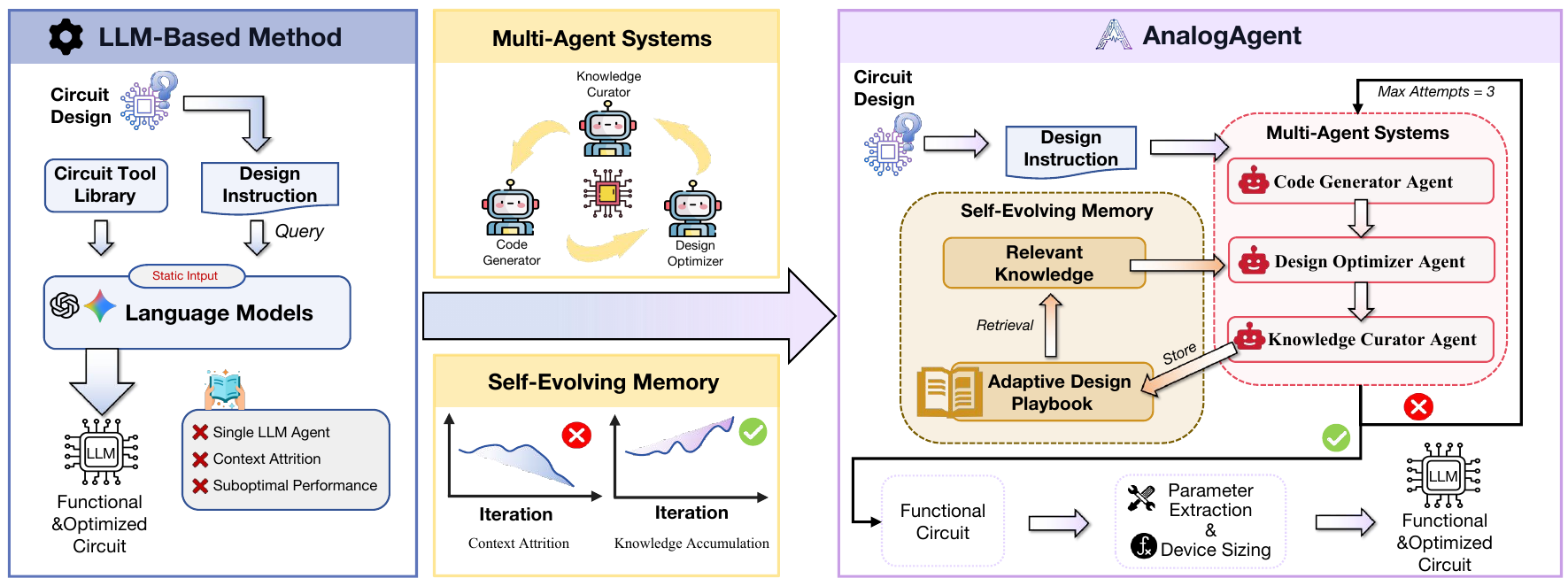}
\vspace{-0.22in}
  \caption{\textbf{AnalogAgent Framework Overview.}  
    \textbf{Left}: Existing LLM-based methods typically rely on a single LLM with static prompts, which often produce suboptimal solutions even when augmented with external circuit libraries. \textbf{Right}: AnalogAgent coordinates multiple agents and curates reusable design knowledge. It retrieves prior knowledge from SEM to refine instructions and perform self-improvement, improving convergence toward functionally complete circuits without additional libraries.}
  \label{fig:demo}
  \vspace{-0.13in}
\end{figure*}

\section{Methodology}

\subsection{Problem Formulation}
We formulate analog circuit design automation as a training-free code generation problem with simulation-based verification. Formally, the input is a task instance $\mathcal{I}=(x,\tau,\Omega,\Phi_{\text{task}})$, where $x$ is the natural-language instruction, $\tau$ is the circuit/task type, $\Omega$ specifies hard interface/API constraints enforced by the evaluation criteria (e.g., required node names, subcircuit pins, and simulator settings), and $\Phi_{\text{task}}=\{\varphi_k\}_{k=1}^K$ is a set of task-specific functional assertions evaluated on simulation outputs. Given $\mathcal{I}$, our AnaglogAgent outputs an executable design candidate $c\in\mathcal{P}$ (SPICE/Python), where $\mathcal{P}$ denotes the space of candidate SPICE/Python programs that satisfy the required interface and API constraints.
Executing $c$ yields $(e,s,z)=\mathcal{E}(c)$, where $e\in\{0,1\}$ indicates program-level or evaluation-criteria results, $s\in\{0,1\}$ indicates simulator-level results, and $z$ denotes the corresponding simulated results and waveforms used to verify design requirements. A candidate design is valid only if all hard constraints and task assertions hold:
\begin{equation}
\label{eq1}
\texttt{PASS}(c)=1 \Longleftrightarrow (e=0)\ \land\ (s=0)\ \land\ \Big(\bigwedge_{k=1}^K \varphi_k(z)=1\Big).
\end{equation}
Violations of $\Omega$ are captured by $e=1$ or $s=1$. In summary, the objective is to map $\mathcal{I}$ to a candidate $c\in\mathcal{P}$ such that $\texttt{PASS}(c)=1$.


\subsection{Framework Overview}
In this paper, we present AnalogAgent, a training-free agentic framework for analog circuit design automation that integrates a multi-agent system (MAS) with self-evolving memory (SEM). As shown in Figure \ref{fig:demo}, the MAS consists of three specialized agents for code generation, circuit design optimization, and knowledge curation, respectively. 
Through iterative self-improvement, the MAS progressively consolidates acquired circuit design knowledge into an Adaptive Design Playbook within the SEM module. 
This mitigates context attrition via structured, incremental updates that preserve fine-grained circuit design knowledge while ensuring contextual representations remain comprehensive during the learning process.
A more detailed workflow is provided in Appendix~\ref{appendixC_workflow}. We detail each component in the following subsections.


\subsection{Multi-agent Systems for Circuit Design}
In analog design tasks, context attrition is particularly acute as the LLM agent must (i) generate executable SPICE/Python code under strict interface constraints across repeated revisions, and (ii) preserve simulation-critical details rather than collapsing them into shorter, less informative summaries that dilute the original instructions and obscure useful guidance.
This motivates the design of MAS that decouples generation, optimization, and knowledge curation into a structured workflow with persistent memory.
MAS maintains a continuously evolving Adaptive Design Playbook that stores distilled, validated heuristics and design constraints. 
Consequently, the Code Generator Agent conditions on task-relevant knowledge retrieved from this playbook, instead of repeatedly reconstructing instructions from transient context.



\noindent\textbf{Code Generator Agent.}
Given a natural-language instruction $x$, the code generator produces task-specific SPICE/Python design scripts. 
Let $t\in\{1,2,\dots\}$ index the iteration. The code generation prompt $P_t$ is constructed by augmenting the task instruction with domain knowledge retrieved from the Adaptive Design Playbook:
\begin{equation}
P_t = P_{\mathrm{task}}(x)\oplus r_t,\qquad
r_t = \mathcal{R}(M_t,\tau),
\end{equation}
where $P_{\mathrm{task}}(x)$ is the task requirements; $r_t$ is the retrieved task-specific knowledge; $\mathcal{R}(\cdot)$ is the retrieval operator; $M_t$ is the SEM state (i.e., the playbook) at iteration $t$; and $\tau$ denotes the circuit/task type.
The code generator then produces a candidate script $c_t$:
\begin{equation}
c_t = \mathcal{G}(P_t),
\end{equation}
where $\mathcal{G}$ denotes the code generator agent.

\noindent\textbf{Design Optimizer Agent.}
After generating the script, a design optimizer agent evaluates it and converts execution feedback into actionable guidance for circuit revision.
Specifically, it performs multi-stage checks: (i) requirement compliance; (ii) DC feasibility via simulation and operating-point inspection; (iii) DC-sweep transfer validation; (iv) task-specific functional tests; and (v) waveform sanity checks \cite{lai2025analogcoder_pro, lai2025analogcoder}. 
Formally, given a candidate script $c_t$, the execution routine returns structured signals:
\begin{equation}
\small
(e_t,\ s_t,\ \ell_t,\ z_t)=\mathcal{E}(c_t),
\end{equation}
where $\mathcal{E}(\cdot)$ denotes the execution and evaluation routine; $e_t\in\{0,1\}$ indicates whether a runtime error occurred at iteration $t$ ($e_t=1$ for failure, $e_t=0$ otherwise); $s_t\in\{0,1\}$ denotes whether a simulator-level failure occurred; $\ell_t$ is the diagnostic log used for feedback; and $z_t$ denotes simulated measurements and waveforms used to verify design requirements. 
A candidate is accepted if Eq.~(\ref{eq1}) evaluates to true, yielding \texttt{PASS}$(c_t)$.

Noted that for circuit classes requiring transfer-curve validation, the design optimizer additionally performs a DC sweep on the current candidate $c_t$ to identify a feasible bias point, and constructs an auxiliary bias-updated candidate for re-evaluation:
\begin{equation}
\small
v_t=\mathcal{B}\!\left(\mathcal{D}(c_t)\right), \qquad
\tilde{c}_t=\mathcal{T}(c_t, v_t),
\end{equation}
where $\mathcal{D}(\cdot)$ produces sweep results for $c_t$, $\mathcal{B}(\cdot)$ selects a feasible bias point from these results, and $\mathcal{T}(\cdot,\cdot)$ injects the selected bias. 

\noindent\textbf{Knowledge Curator Agent.}
 Inspired by recent agentic frameworks that store salient contextual information for accumulate, retrieve, and reason over long-horizon experiences \cite{suzgun2025dynamic, zhang2025agentic, agrawal2025gepa}, we propose a Knowledge Curator that consolidates the design optimizer’s feedback into SEM by maintaining and refining an Adaptive Design Playbook. 
Rather than rewriting the full context, it performs incremental updates that extract compact, reusable rules and code-level patterns from both failures and successes. 
This preserves validated guidance while incorporating task-driven insights, mitigating context attrition and enabling reliable retrieval of domain-specific tactics for reference.
Formally, given a validated feedback bundle, the curator distills reusable knowledge and exemplars conditioned on the circuit/task type $\tau$, which are then incorporated into SEM through incremental memory updates across iterations. 
\begin{equation}
f_t=\bigl(x,c_t,e_t,s_t,\ell_t, z_t, \texttt{PASS}(c_t)\bigr), \Delta_t=\phi(f_t),
\end{equation}
where $f_t$ aggregates the feedback information and $\phi(\cdot)$ maps this feedback to structured circuit design knowledge $\Delta_t$, which is stored in the memory mechanism described in the following subsection.




\subsection{Self-Evolving Memory Mechanism}
LLMs benefit from long, detailed circuit-design instructions, yet as instruction complexity and iteration depth grow, LLMs exhibit a tendency to compress contexts into distilled summaries, which accelerates context attrition. 
This motivates the introduction of self-evolving memory (SEM), maintained as an Adaptive Design Playbook that incrementally refines structured reusable circuit design knowledge. 
Specifically, SEM constitutes a training-free, dynamically updated repository that facilitates cross-task generalization and enhances convergence on complex circuit design tasks. In doing so, it reduces reliance on repeated full-context rewrites and mitigates information loss across iterative self-improvement, while operating without expert feedback or reliance on predefined circuit libraries or databases. 
Overall, by continuously updating the Adaptive Design Playbook and retrieving task-relevant design knowledge, MAS and SEM establish a closed-loop self-improvement paradigm for analog circuit design automation.



\noindent\textbf{Adaptive Design Playbook Update.} 
AnalogAgent maintains a SEM repository as an Adaptive Design Playbook that accumulates validated rules and exemplars across iterations through incremental, curator-approved updates. 
Given the circuit design knowledge extracted by the Knowledge Curator Agent, the playbook is continuously updated as follows:
\begin{equation}
M_{t+1} = \mathcal{U}\left(M_t,\ \Delta_t\right)
=
\mathrm{Dedup}\big(
\mathrm{Filter}(M_t \mathbin{\Vert} \Delta_t;\Omega)
\big),
\end{equation}
where $\mathcal{U}(\cdot)$ updates the playbook $M_t$ with conflict checking under interface/API constraints $\Omega$, followed by filtering, deduplication process to suppress erroneous experience updates, eliminate redundant entries, and maintain a compact, structured memory.


\noindent\textbf{Design Knowledge Retrieval.}
When encountering a new circuit design task, relevant knowledge is retrieved from the playbook to guide script generation, thereby improving both consistency and efficiency.
Given the current task type $\tau$ (and optional recent failure reflection $\mathcal{F}$), a retriever first queries task-specific memory in $M_t$ by type match and resorts to substring-based retrieval when such a match is unavailable, and then forms
\begin{equation}
r(\tau)=
\begin{cases}
\mathcal{R}[\tau], & \tau\in\mathrm{keys}(r),\\
\mathcal{R}[k^{*}],  & k^{*}=\min\{k\in\mathrm{keys}(r) : k\prec \mathrm{str}(\tau)\},\\
\varnothing,         & \text{otherwise}.
\end{cases}
\end{equation}
\noindent where $k\prec \mathrm{str}(\tau_t)$ denotes a substring match i.e., $k$ is contained within the string representation of $\tau$. 
This prioritizes exact task-aligned matches, while enabling approximate reuse of related entries when exact matches are unavailable, thereby improving robustness and coverage.


\noindent\textbf{Closed-loop Self-improvement.}
During the self-improving process, AnalogAgent forms a closed loop in which the updated memory state $M_{t+1}$ is retrieved to construct the retrieved knowledge $r_{t+1}$, which is then combined with the task instruction to form $P_{t+1}$ for generating the next candidate $c_{t+1}$:
\begin{equation}
\begin{aligned}
r_{t+1}&=\mathcal{R}(M_{t+1},\tau),\\P_{t+1}&=P_{\mathrm{task}}(x)\oplus r_{t+1},\\
c_{t+1}&=\mathcal{G}(P_{t+1}).
\end{aligned}
\end{equation}
This completes the loop between generation, verification, and knowledge updates and retrieval. In summary, MAS provides the operational mechanism for producing and applying structured feedback for circuit design knowledge curation, while SEM serves as persistent memory that preserves these updates and supports self-improving across iterations and cross-task knowledge transfer.

\subsection{Circuit Optimization}
After a feasible circuit design is generated, the design is further refined through parameter optimization. 
Following AnalogCoder-Pro \cite{lai2025analogcoder_pro}, the circuit is first validated via simulation to ensure executability. 
Based on the task instruction $x$ and simulated results $z_t$, the LLM identifies key tunable variables (e.g., device dimensions or component values) together with their corresponding search ranges. 
Bayesian optimization using the Tree-structured Parzen Estimator (TPE) implemented in Optuna \cite{frazier2018tutorial} is then employed to iteratively evaluate candidate parameter configurations through simulation and select the configuration that satisfies the target specifications.
\section{Experiments}
\subsection{Experimental Settings}
\label{sec:exp_settings}
\textbf{Benchmark.} 
We evaluate on a 30-task analog circuit benchmark grouped by component count and topological complexity (Tasks~1--8 \emph{Easy}, Tasks~9--13 \emph{Medium}, Tasks~14--30 \emph{Hard}), which is publicly available \footnote{https://github.com/laiyao1/AnalogCoder} and adopted from prior work \cite{lai2025analogcoder, lai2025analogcoder_pro}. 
We provide clear, detailed descriptions of how evaluation data are collected through SPICE/Python execution (e.g., simulation logs, measurements, and waveforms), how these artifacts are preprocessed and managed across runs, and how results are analyzed under a unified protocol to ensure rigorous, reproducible assessment. Please refer to Appendix~\ref{appendixB} for more details.

\noindent\textbf{Baselines.} We compare AnalogAgent with \textbf{SPICEPilot} \cite{vungarala2024spicepilot}, \textbf{AnalogCoder} \cite{lai2025analogcoder}, and \textbf{AnalogCoder-Pro} \cite{lai2025analogcoder_pro}. SPICEPilot uses prompt-based generation with curated instructions and in-context examples \cite{vungarala2024spicepilot}. AnalogCoder introduces a Circuit Tool Library for reusable components and structured synthesis \cite{lai2025analogcoder}, and AnalogCoder-Pro further adds a multimodal feedback-enhanced design flow \cite{lai2025analogcoder_pro}. 
The evaluation includes a diverse set of LLMs, including DeepSeek-V2-Lite, Llama-3-8B, Gemini-2.5-Flash, GPT-5. 
All methods use identical configurations for fair comparisons, with full experimental configurations provided in Appendix~\ref{appendixB}.

\noindent\textbf{Metrics.} 
We follow prior work  \cite{lai2025analogcoder,lai2025analogcoder_pro} and use Pass@$k$ to assess the reliability of generated circuit designs, defined as the probability of obtaining at least one valid solution among $k$ samples:
$\text{Pass@k} = 1 - \frac{\binom{n-c}{k}}{\binom{n}{k}}$,
where $n$ is the total number of candidates and $c$ is the number of correct solutions. Throughout, we fix $n=30$ to obtain stable estimates under stochastic model outputs.

\subsection{Experimental Results}

Table~\ref{tab:main-result} reports primary results on 30 analog circuit design tasks under a unified benchmark and evaluation protocol, spanning multiple open-source LLMs and LLM baselines. The open-source DeepSeek-V2-Lite achieves Pass@1/Pass@5 of 6.7/11.4, while Llama3-8B attains 8.6/27.0. 
SPICEPilot exhibits relatively low Pass@1 (50.1) but much higher Pass@5 (96.5), indicating that it often reaches a valid executable solution only after multiple independent samples rather than in the first attempt. 
Under GPT-5, AnalogCoder performs poorly on Hard tasks (Pass@1 36.5), while AnalogCoder-Pro improves the average Pass@1/Pass@5 to 88.6/98.2 but remains limited on several high-difficulty cases (e.g., Tasks 24–25). On the other hand, AnalogAgent consistently attains the best overall Pass@1 and Pass@5 among all methods. Specifically, with GPT-5 backbone, AnalogAgent achieves 97.4/100.0 Pass@1/Pass@5 and improves Hard-task Pass@1 over AnalogCoder-Pro by around 12.5\%. With Gemini-2.5-Flash, AnalogAgent reaches 92.0/99.9, suggesting a lower first-try success rate but a comparable multi-sample ceiling. 

Table~\ref{tab:5times_comparison} complements these aggregate results with statistics over five random trials, reporting Pass@1 as mean±std for Easy/Medium/Hard groups. AnalogAgent achieves the highest mean Pass@1 with smaller variance across trials, including $\mathbf{87.1 \pm 2.9}$ on Medium and $\mathbf{88.7 \pm 0.9}$ on Hard tasks, indicating more stable first-attempt performance in the regimes where prior methods struggle. Beyond success rates, Table~\ref{tab:5times_comparison} also summarizes efficiency using the average tokens and runtime required to reach the first success across all 30 tasks. AnalogAgent reduces Time-to-First-Success to 1.3 minutes on average (vs. 2.6 for SPICEPilot and 2.1 for AnalogCoder-Pro) and lowers tokens to first success to $(\sim)32K (vs. (\sim)40K$ for AnalogCoder-Pro). Taken together, these results indicate that AnalogAgent’s gains are not driven by increased search effort; instead, by structuring reusable knowledge into an Adaptive Design Playbook and retrieving targeted guidance in subsequent iterations, our framework improves both first-shot reliability and convergence efficiency relative to existing baselines.

\newcommand{\score}[1]{%
  \IfStrEq{#1}{100.0}{\textbf{#1}}{#1}%
}
\begin{table*}[!t]
\centering
\caption{\textbf{Main results.} Across diverse LLM backbones and model variants under a unified benchmark and evaluation protocol, reporting per-task and total tasks Pass@1 and Pass@5.}
\vspace{-0.1in}
\label{tab:main-result}
\scriptsize
\setlength{\tabcolsep}{2.2pt}
\renewcommand{\arraystretch}{1.05}
\renewcommand{\score}[1]{#1}
\resizebox{\textwidth}{!}{%
\begin{tabular}{c cc cc cc cc cc cc cc|cc cc}
\toprule
\textbf{Model}
& \multicolumn{2}{c}{\textbf{DeepSeek-V2-Lite}}
& \multicolumn{2}{c}{\textbf{Llama3-8B}}
& \multicolumn{2}{c}{\makecell{\textbf{SPICEPilot}\\\textbf{(Gemini-2.5-Flash)}}}
& \multicolumn{2}{c}{\makecell{\textbf{AnalogCoder}\\\textbf{(Gemini-2.5-Flash)}}}
& \multicolumn{2}{c}{\makecell{\textbf{AnalogCoder}\\\textbf{(GPT-5)}}}
& \multicolumn{2}{c}{\makecell{\textbf{AnalogCoder-Pro}\\\textbf{(Gemini-2.5-Flash)}}}
& \multicolumn{2}{c|}{\makecell{\textbf{AnalogCoder-Pro}\\\textbf{(GPT-5)}}}
& \multicolumn{2}{c}{\makecell{\textbf{AnalogAgent}\\\textbf{(Gemini-2.5-Flash)}}}
& \multicolumn{2}{c}{\makecell{\textbf{AnalogAgent}\\\textbf{(GPT-5)}}} \\
\textbf{Task ID}
& \textbf{Pass@1} & \textbf{Pass@5}
& \textbf{Pass@1} & \textbf{Pass@5}
& \textbf{Pass@1} & \textbf{Pass@5}
& \textbf{Pass@1} & \textbf{Pass@5}
& \textbf{Pass@1} & \textbf{Pass@5}
& \textbf{Pass@1} & \textbf{Pass@5}
& \textbf{Pass@1} & \textbf{Pass@5}
& \textbf{Pass@1} & \textbf{Pass@5}
& \textbf{Pass@1} & \textbf{Pass@5} \\
\midrule
1  & \score{83.3} & \score{100.0} & \score{26.7} & \score{81.5} & \score{60.0} & \score{99.4} & \score{73.3} & \score{100.0} & \score{100.0} & \score{100.0} & \score{100.0} & \score{100.0} & \score{100.0} & \score{100.0} & \score{100.0} & \score{100.0} & \score{100.0} & \score{100.0} \\
2  & \score{0.0}  & \score{0.0}   & \score{30.0} & \score{85.7}  & \score{63.3} & \score{99.7} & \score{70.0} & \score{99.9} & \score{100.0} & \score{100.0} & \score{100.0} & \score{100.0} & \score{100.0} & \score{100.0} & \score{100.0}  & \score{100.0} & \score{100.0} & \score{100.0} \\
3  & \score{0.0}  & \score{0.0}   & \score{20.0} & \score{70.2} & \score{66.7} & \score{99.8}& \score{90.0} & \score{100.0} & \score{100.0} & \score{100.0} & \score{100.0} & \score{100.0} & \score{100.0} & \score{100.0} & \score{100.0} & \score{100.0} & \score{100.0} & \score{100.0} \\
4  & \score{0.0}  & \score{0.0}   & \score{3.3}  & \score{16.7} & \score{60.0} & \score{99.4} & \score{73.3} & \score{100.0} & \score{100.0} & \score{100.0} & \score{80.0}  & \score{100.0} & \score{100.0} & \score{100.0} & \score{100.0} & \score{100.0} & \score{100.0} & \score{100.0} \\
5  & \score{0.0}  & \score{0.0}   & \score{0.0}  & \score{0.0} & \score{63.3} & \score{99.7} & \score{80.0} & \score{100.0} & \score{100.0} & \score{100.0} & \score{93.3}  & \score{100.0} & \score{100.0} & \score{100.0} & \score{100.0} & \score{100.0} & \score{100.0}  & \score{100.0} \\
6  & \score{13.3} & \score{53.8}  & \score{26.7} & \score{81.5} & \score{60.0} & \score{99.4}& \score{100.0}& \score{100.0} & \score{100.0} & \score{100.0} & \score{100.0} & \score{100.0} & \score{100.0} & \score{100.0} & \score{100.0} & \score{100.0} & \score{100.0} & \score{100.0} \\
7  & \score{0.0}  & \score{0.0}   & \score{30.0} & \score{85.7} & \score{43.3} & \score{95.7} & \score{100.0}& \score{100.0} & \score{100.0} & \score{100.0} & \score{100.0} & \score{100.0} & \score{100.0} & \score{100.0} & \score{100.0} & \score{100.0} & \score{100.0} & \score{100.0} \\
8  & \score{3.3}  & \score{16.7}  & \score{13.3} & \score{53.8} & \score{40.0} & \score{94.0} & \score{80.0} & \score{100.0} & \score{76.7} & \score{100.0} & \score{76.7}  & \score{100.0} & \score{100.0} & \score{100.0} & \score{100.0} & \score{100.0} & \score{100.0} & \score{100.0} \\
9  & \score{20.0} & \score{70.2}  & \score{0.0}  & \score{0.0} & \score{50.0} & \score{97.9}  & \score{10.0} & \score{43.3}  & \score{96.7} & \score{100.0} & \score{26.7}  & \score{81.5}  & \score{100.0} & \score{100.0} & \score{66.7}  & \score{99.8}  & \score{100.0} & \score{100.0} \\
10 & \score{0.0}  & \score{0.0}   & \score{23.3} & \score{76.4} & \score{56.7} & \score{99.1} & \score{96.7} & \score{100.0} & \score{100.0} & \score{100.0} & \score{100.0} & \score{100.0} & \score{100.0} & \score{100.0} & \score{100.0} & \score{100.0} & \score{100.0} & \score{100.0} \\
11 & \score{0.0}  & \score{0.0}   & \score{0.0}  & \score{0.0}  & \score{33.3} & \score{89.1} & \score{16.7} & \score{62.7}  & \score{100.0} & \score{100.0} & \score{100.0} & \score{100.0} & \score{100.0}  & \score{100.0} & \score{100.0} & \score{100.0} & \score{100.0} & \score{100.0} \\
12 & \score{0.0}  & \score{0.0}   & \score{0.0}  & \score{0.0} & \score{46.7} & \score{96.9}  & \score{23.3} & \score{76.4}  & \score{43.3}  & \score{95.7}  & \score{30.0}  & \score{85.7}  & \score{66.7}  & \score{99.8}  & \score{70.0}  & \score{99.9} & \score{100.0} & \score{100.0} \\
13 & \score{0.0}  & \score{0.0}   & \score{26.7} & \score{81.5}& \score{53.3} & \score{98.6}  & \score{33.3} & \score{89.1}  & \score{86.7} & \score{100.0} & \score{96.7}  & \score{100.0} & \score{100.0} & \score{100.0} & \score{96.7}  & \score{100.0} & \score{100.0} & \score{100.0} \\
14 & \score{80.0} & \score{100.0} & \score{33.3} & \score{89.1} & \score{63.3} & \score{99.7} & \score{83.3} & \score{100.0} & \score{93.3}  & \score{100.0} & \score{100.0} & \score{100.0} & \score{100.0} & \score{100.0} & \score{100.0} & \score{100.0} & \score{100.0} & \score{100.0} \\
15 & \score{0.0}  & \score{0.0}   & \score{20.0} & \score{70.2} & \score{46.7} & \score{96.9} & \score{23.3} & \score{76.4}  & \score{50.0}  & \score{97.9}  & \score{90.0}  & \score{100.0} & \score{56.7}  & \score{99.1}  & \score{100.0} & \score{100.0} & \score{100.0} & \score{100.0} \\
16 & \score{0.0}  & \score{0.0}   & \score{0.0}  & \score{0.0}  & \score{53.3} & \score{98.6} & \score{0.0}  & \score{0.0}   & \score{23.3}   & \score{76.4}  & \score{66.7}  & \score{99.8}  & \score{16.7}  & \score{62.7}  & \score{70.0}  & \score{99.9}  & \score{83.3}  & \score{100.0} \\
17 & \score{0.0}  & \score{0.0}   & \score{3.3}  & \score{16.7} & \score{43.3} & \score{95.7} & \score{0.0}  & \score{0.0}   & \score{0.0}   & \score{0.0}   & \score{76.7}  & \score{100.0} & \score{100.0} & \score{100.0} & \score{86.7}  & \score{100.0} & \score{90.0}  & \score{100.0} \\
18 & \score{0.0}  & \score{0.0}   & \score{0.0}  & \score{0.0} & \score{63.3} & \score{99.7}  & \score{93.3}  & \score{100.0}   & \score{90.0} & \score{100.0} & \score{100.0} & \score{100.0} & \score{100.0} & \score{100.0} & \score{100.0} & \score{100.0} & \score{100.0} & \score{100.0} \\
19 & \score{0.0}  & \score{0.0}   & \score{0.0}  & \score{0.0} & \score{56.7} & \score{99.1}  & \score{0.0}  & \score{0.0}   & \score{0.0}  & \score{0.0}  & \score{100.0} & \score{100.0} & \score{100.0} & \score{100.0} & \score{100.0} & \score{100.0} & \score{100.0} & \score{100.0} \\
20 & \score{0.0}  & \score{0.0}   & \score{0.0}  & \score{0.0} & \score{53.3} & \score{98.6}  & \score{0.0}  & \score{0.0}   & \score{0.0} & \score{0.0} & \score{100.0} & \score{100.0} & \score{80.0}  & \score{100.0} & \score{100.0} & \score{100.0} & \score{83.3}  & \score{100.0} \\
21 & \score{0.0}  & \score{0.0}   & \score{0.0}  & \score{0.0} & \score{36.7} & \score{91.8}  & \score{0.0}  & \score{0.0}   & \score{0.0}  & \score{0.0}  & \score{93.3}  & \score{100.0} & \score{96.7}  & \score{100.0} & \score{100.0} & \score{100.0} & \score{100.0} & \score{100.0} \\
22 & \score{0.0}  & \score{0.0}   & \score{0.0}  & \score{0.0} & \score{33.3} & \score{89.1}  & \score{0.0}  & \score{0.0}   & \score{0.0}   & \score{0.0}   & \score{100.0} & \score{100.0} & \score{90.0}  & \score{100.0} & \score{100.0} & \score{100.0} & \score{86.7}  & \score{100.0} \\
23 & \score{0.0}  & \score{0.0}   & \score{0.0}  & \score{0.0}  & \score{50.0} & \score{97.9} & \score{90.0} & \score{100.0} & \score{0.0}   & \score{0.0}   & \score{83.3}  & \score{100.0} & \score{80.0}    & \score{100.0}    & \score{73.3}  & \score{100.0} & \score{100.0} & \score{100.0} \\
24 & \score{0.0}  & \score{0.0}   & \score{0.0}  & \score{0.0} & \score{43.3} & \score{95.7}  & \score{16.7} & \score{62.7}  & \score{0.0}   & \score{0.0}   & \score{33.3}  & \score{89.1}  & \score{36.7}    & \score{91.8}    & \score{86.7}  & \score{100.0} & \score{100.0} & \score{100.0} \\
25 & \score{0.0}  & \score{0.0}   & \score{0.0}  & \score{0.0} & \score{30.0} & \score{85.7}  & \score{26.7} & \score{81.5}  & \score{0.0}   & \score{0.0}   & \score{3.3}   & \score{16.7}  & \score{40.0}  & \score{94.0}  & \score{53.3}  & \score{98.6}  & \score{86.7}  & \score{100.0} \\
26 & \score{0.0}  & \score{0.0}   & \score{0.0}  & \score{0.0} & \score{33.3} & \score{89.1}  & \score{83.3} & \score{100.0} & \score{100.0}  & \score{100.0}  & \score{96.7}  & \score{100.0} & \score{100.0} & \score{100.0} & \score{100.0} & \score{100.0} & \score{100.0}  & \score{100.0} \\
27 & \score{0.0}  & \score{0.0}   & \score{0.0}  & \score{0.0} & \score{50.0}  & \score{97.9}  & \score{90.0} & \score{100.0} & \score{100.0}  & \score{100.0} & \score{100.0} & \score{100.0} & \score{100.0} & \score{100.0} & \score{100.0} & \score{100.0} & \score{93.3}  & \score{100.0} \\
28 & \score{0.0}  & \score{0.0}   & \score{0.0}  & \score{0.0}  & \score{36.7}  & \score{91.8} & \score{0.0}  & \score{0.0}   & \score{33.3}  & \score{89.1}  & \score{100.0} & \score{100.0} & \score{96.7}  & \score{100.0} & \score{100.0} & \score{100.0} & \score{100.0} & \score{100.0} \\
29 & \score{0.0}  & \score{0.0}   & \score{0.0}  & \score{0.0} & \score{56.7}  & \score{99.1}  & \score{73.3} & \score{100.0} & \score{100.0}  & \score{100.0}  & \score{100.0} & \score{100.0} & \score{100.0} & \score{100.0} & \score{96.7}  & \score{100.0} & \score{100.0} & \score{100.0} \\
30 & \score{0.0}  & \score{0.0}   & \score{0.0}  & \score{0.0} & \score{56.7}  & \score{99.1}  & \score{0.0} & \score{0.0} & \score{30.0}  & \score{85.7}  & \score{3.3}   & \score{16.7}  & \score{96.7}  & \score{100.0} & \score{60.0}  & \score{99.8} & \score{100.0} & \score{100.0} \\

\midrule
Avg
& \score{6.7} & \score{11.4}
& \score{8.6} & \score{27.0}
& \score{50.1} & \score{96.5}
& \score{47.6} & \score{66.4}
& \score{60.8} & \score{71.5}
& \score{81.7} & \score{93.0}
& \score{88.6} & \score{98.2}
& \score{92.0} & \score{99.9}
& \score{\textbf{97.4}} & \score{\textbf{100.0}} \\
\midrule
Imp
& \score{1362.4} & \score{780.5}
& \score{1039.2} & \score{270.8}
& \score{94.5} & \score{3.7}
& \score{104.9} & \score{50.6}
& \score{60.3} & \score{39.9}
& \score{19.3} & \score{7.5}
& \score{10.0} & \score{1.8}
& \score{5.9} & \score{0.1}
& \score{0} & \score{0} \\
\bottomrule
\end{tabular}%
}
\vspace{0.15in}
\end{table*}

\begin{table}[!ht]
\centering
\scriptsize
\setlength{\tabcolsep}{4.5pt}
\renewcommand{\arraystretch}{1.5}
\caption{
Average Pass@1 success rates by task difficulty and average token (K) /runtime (min) cost to first success (mean$\pm$std over five trials). All methods use Gemini-2.5-Flash for fairness; NA denotes at least one task unsolved, rendering the aggregate undefined.}
\label{tab:5times_comparison}
\resizebox{\columnwidth}{!}{%
\begin{tabular}{c|cccccc}
\hline\hline
\makecell{\textbf{Model}} 
    & \textbf{Easy} & \textbf{Medium} & \textbf{Hard} & \textbf{Avg Results}  & \textbf{Avg Tokens}& \textbf{Avg Time}\\
\hline
\textbf{DeepSeek-V2-Lite} 
& $12.7 \pm 2.91$ & $10.0 \pm 2.9$ & $3.4 \pm 1.3$ & $6.4 \pm 2.5$ &NA&NA\\
\textbf{Llama3-8B} 
& $18.9 \pm 11.9$ & $10.1 \pm 5.6$ & $1.8 \pm 1.2$ & $7.8 \pm 4.5$ & NA&NA\\
\textbf{SPICEPilot} 
& $48.4 \pm 15.0$ & $47.1 \pm 13.1$ & $40.7 \pm 6.3$ & $43.7 \pm 9.3$ & 251.5&2.6\\
\makecell{\textbf{AnalogCoder}} 
& $87.6 \pm 4.8$ & $34.0 \pm 4.2$ & $39.1 \pm 1.9$ & $50.9 \pm 2.6$ &NA&NA\\
\makecell{\textbf{AnalogCoder-Pro}} 
& $91.9 \pm 3.4$ & $83.2 \pm 4.4$ & $73.3 \pm 1.7$ & $81.7 \pm 1.1$ &40.2&2.1\\
\makecell{\textbf{AnalogAgent (Ours)}} 
& $\mathbf{98.3 \pm 1.6}$ & $\mathbf{87.1 \pm 2.9}$ & $\mathbf{88.7 \pm 0.9}$ & $\mathbf{92.3 \pm 0.6}$ & \textbf{32.0}&\textbf{1.3}\\
\hline\hline
\end{tabular}%
}
\end{table}
\vspace{-0.15in}
\begin{table*}[t]
\centering
\caption{\textbf{Ablation study.} A series of ablation studies on the AnalogAgent built on Gemini-2.5-Flash validate the effectiveness of the proposed method by systematically removing individual components of the framework.}
\label{tab:ablation}
\tiny
\setlength{\tabcolsep}{5pt}
\renewcommand{\arraystretch}{1.12}

\resizebox{\linewidth}{!}{%
\begin{tabular}{c cc cc cc cc cc}
\toprule
\textbf{Model}&
\multicolumn{2}{c}{\textbf{Gemini-2.5-Flash}} &
\multicolumn{2}{c}{\makecell{\textbf{AnalogAgent}\\\textbf{(w/o MAS, w/o SEM)}}} &
\multicolumn{2}{c}{\makecell{\textbf{AnalogAgent}\\\textbf{(w/ MAS, w/o SEM)}}} &
\multicolumn{2}{c}{\makecell{\textbf{AnalogAgent}\\\textbf{(w/o MAS, w/ SEM)}}} &
\multicolumn{2}{c}{\textbf{AnalogAgent}} \\
Task Level& Pass@1 & Pass@5 & Pass@1 & Pass@5 & Pass@1 & Pass@5 & Pass@1 & Pass@5 & Pass@1 & Pass@5 \\
\midrule
Easy     & 78.3 & 85.2 & 82.9 & 97.7 & 94.2 & 100.0 & 88.8 & 98.6 & 100.0 & 100.0 \\
Medium & 60.0 & 60.0 & 64.7 & 80.0 & 75.3 & 92.5  & 56.0 & 66.2 & 86.7 & 99.6  \\
Hard  & 34.5 & 41.0 & 43.7 & 52.8 & 68.8 & 70.9  & 45.1 & 54.3 & 89.8 & 100.0 \\
\midrule
Avg            & 50.4 & 56.0 & 63.8 & 76.8 & 79.4 & 87.8  & 58.6 & 68.1 & \textbf{92.0} & \textbf{99.9} \\

\bottomrule
\end{tabular}%
}
\vspace{-0.05in}
\end{table*}

\subsection{Ablation Studies}
\label{sec:ablation}
To assess the contributions of SEM and MAS, we conduct ablations with Gemini-2.5-Flash under four configurations: (i) the baseline Gemini without agentic famework AnalogAgent; MAS only retains multi-agent system with iteration-level feedback exchange, but disables curator-driven memory updates and playbook retrieval; SEM only enables curator-style rule distillation with playbook-guided generation, but removes multi-agent and operates in a single-agent loop.
As summarized in Table ~\ref{tab:ablation}, the removal of both modules results in a clear reduction in the success of analog circuit design automation, with the most pronounced degradation observed in high-complexity tasks. On Hard tasks, Pass@1 decreases from 89.8 in the full framework to 43.7 when both modules are removed, and the overall Avg Pass@1/Pass@5 drops from 92.0/99.9 to 63.8/76.8. The SEM is associated with higher performance on Easy tasks by adding revelant knowledge. Easy-task Pass@1 improves from 78.3 to 88.8 with SEM only and reaches 100.0 in the full model. In contrast, MAS shows a larger effect on Medium and Hard tasks, where multi-agent design patterns assist in resolving complex topological constraints and bias relationships. Medium-task Pass@1 increases from 60.0 to 75.3 with MAS only, while Hard-task Pass@1 increases from 34.5 to 68.8. The full framework yields the strongest results among all variants, demonstrating that short-term refinement and long-term information reuse are jointly necessary and most effective when combined.

\captionsetup{labelfont=normalfont,textfont=normalfont}
\begin{figure*}[t]
  \centering
  \includegraphics[width=1\linewidth]{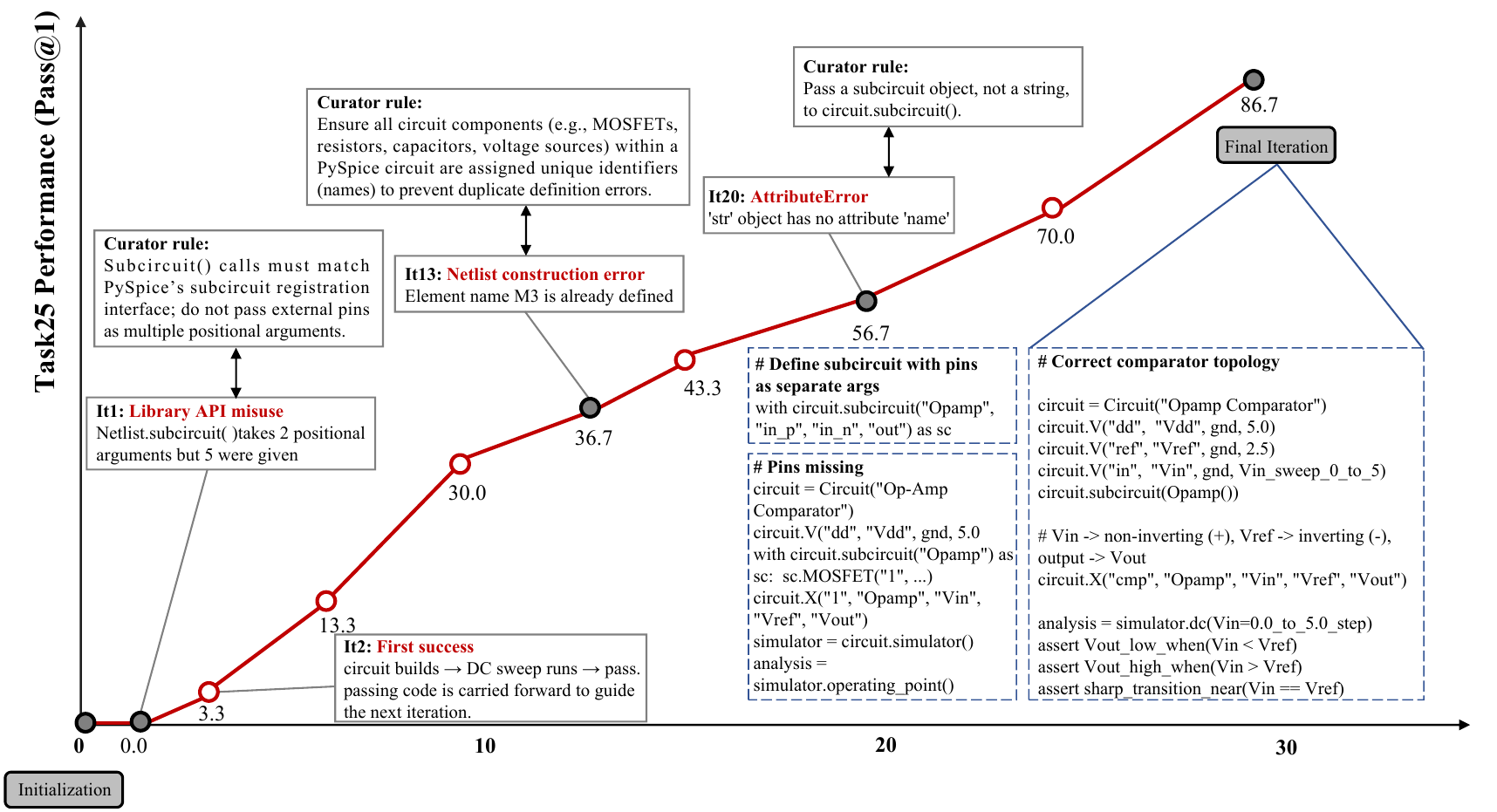}
  \caption{\textbf{Self-Improving Refinement Case Study.} The flow shows an error-driven refinement loop for Hard Task 25, in which execution failures are diagnosed, distilled into curator rules, and written into Adaptive Design Playbook to guide subsequent generations and accelerate convergence. Hard Task 25 is an op-amp comparator that outputs high when $V_{\text{in}} > V_{\text{ref}}$ and low when $V_{\text{in}} < V_{\text{ref}}$. The final iterations correct the topology and pin semantics to meet this specification.}
\label{fig:Example}
\end{figure*}
\captionsetup{labelfont=normalfont,textfont=normalfont}
\begin{figure*}[t]
  \centering
  \includegraphics[width=1.0\linewidth]{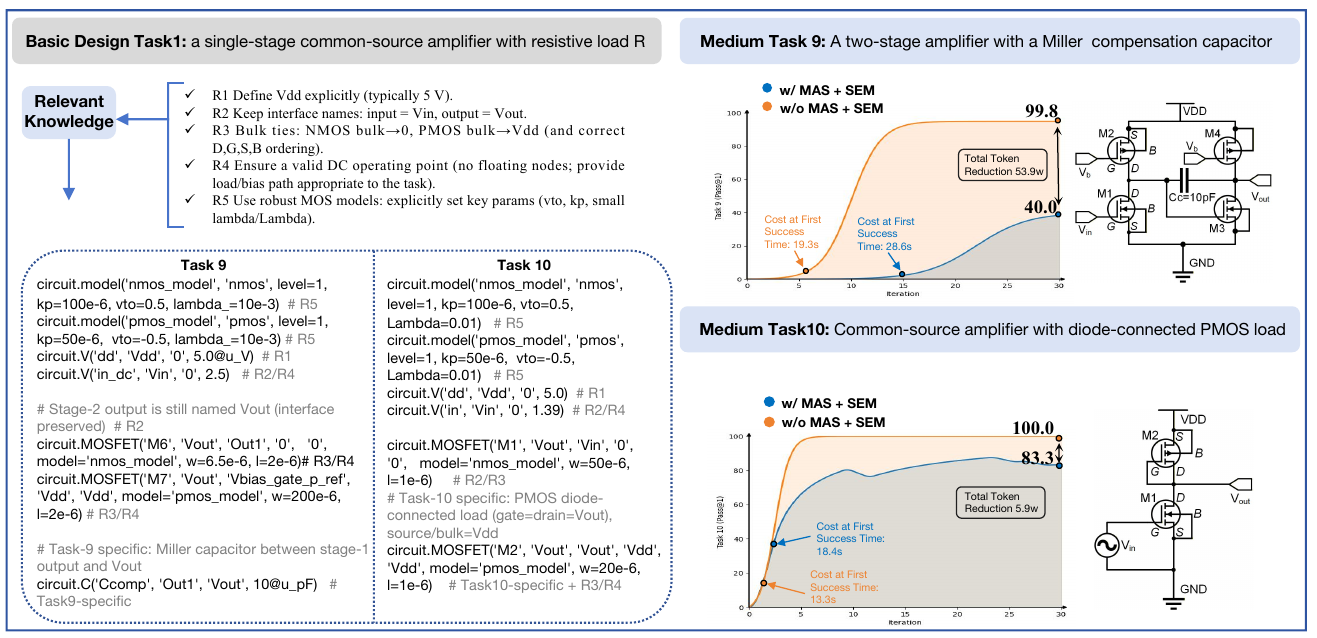}
  \caption{\textbf{Cross-Task Transfer across Amplifiers.} Reusable knowledge distilled from the basic Amplifier Task 1 transfers to Tasks 9 and Task 10, improving success rates and demonstrating that long-term memory is effectively reusable across tasks.
}
\label{fig:cross}
\end{figure*}

\subsection{Self-Improving Refinement Case Study}
\label{sec:visualization_case}

To better understand the behavior of the proposed framework during circuit synthesis, this section analyzes representative design trajectories and iterative correction patterns. Task~25, an Op-amp Comparator, is selected as an illustrative example to demonstrate the progression from initial failure to a valid solution. As shown in Figure \ref{fig:Example}, the process spans 30 iterations, during which successive generations are refined based on execution feedback. Early attempts exhibit library usage errors and netlist construction issues, such as duplicated element definitions. In contrast to baseline frameworks that frequently repeat similar syntax-level mistakes, AnalogAgent shows a progressive reduction of such errors across iterations. Execution feedback highlights concrete problems, which are recorded by the Knowledge Curator as reusable correction rules, including ensuring unique identifiers for MOSFET instances and passing subcircuit objects rather than string labels. In later iterations, the generated topology and pin connections conform to the comparator specification, with $V_{out}$ switching behavior aligned with the condition $V_{in}=V_{ref}$. Notably, AnalogAgent attains a final Pass@1 of 86.7 on Task~25, representing a substantial accuracy improvement over other frameworks on this high-difficulty case.

Throughout the iteration, AnalogAgent first learns the exact number and ordering of positional arguments needed to instantiate each subcircuit topology. When generating the corresponding netlist, it enforces globally unique instance identifiers within each scope to avoid name collisions and to ensure a one-to-one mapping between devices and their stamped mathematical objects during SPICE equation assembly. Building on these rules, AnalogAgent then generates a hierarchical comparator: it defines the “Opamp” subcircuit with a consistent pin order, and then instantiates it at the top level with correctly named, fully connected pins, including supplies and references, yielding a SPICE-simulatable circuit.

\subsection{Case Study: Cross-Task Knowledge Transfer}
The cumulative experience further supports convergence on related tasks through cross-task reuse of previously recorded design information. For example, information derived from Basic Task~1 regarding explicit supply definition and bulk-tie conventions is retrieved by the Adaptive Design Playbook when addressing Medium Tasks~9 and~10. As illustrated in Figure~\ref{fig:cross}, such reuse preserves consistent biasing strategies and interface naming conventions across tasks, thereby narrowing the range of candidate designs explored for more complex circuits. By incorporating feedback from unsuccessful attempts into updated Playbook entries, the framework replaces isolated sampling with a process guided by stored diagnostic information.

During the generation for Task~1, AnalogAgent learns and retains the circuit-construction grammar required for SPICE netlisting. Concretely, it internalizes the physical pin ordering and connectivity rules for each primitive; for instance, a MOSFET instance is specified in the order of unique identifier, drain, gate, source, body, model name, length, and width, while passive elements require valid two-terminal node connections. It also learns the need for unique instance names (e.g., M1, R1, Vdd) to avoid identifier collisions during parsing and device stamping in simulation. With this foundational ``syntax + wiring'' knowledge established in Task~1, which implements a simple NMOS pull-down with a resistive pull-up and a safety resistor, the agent transfers the same rules to more complex topologies in Task~9 and Task~10.
For Task~9, it reuses the MOSFET netlisting template and expands from a single-stage inverter-like behavior to a two-stage biased CMOS structure by introducing intermediate nodes (Vint1), bias sources (Vbias1, Vbias2), complementary PMOS loads, and a coupling capacitor (Cc), while preserving correct pin order, valid node naming, and model references. For Task~10, it again applies the same instantiation rule-set to generate a regenerative/inverter-style topology (NMOS pull-down plus diode-connected or feedback PMOS load), ensuring the feedback connection, where M2 gate is tied to Vout, remains physically meaningful and simulatable. Overall, AnalogAgent’s workflow reflects a progressive transfer from syntax correctness to connectivity consistency and then to topology composition: it first learns how to write valid device lines and assign instance names in Task~1, and subsequently reuses this knowledge to reduce trial-and-error iterations when assembling larger hierarchical or multi-device circuits in Tasks~9 and~10, thereby focusing the remaining search on architectural choices such as biasing, intermediate nodes, and feedback rather than re-learning netlist formatting from scratch.


\subsection{Agentic Reasoning with "Small" LLMs}
\label{sec:small_models}
\begin{figure}[!t]
  \centering
  \includegraphics[width=1.0\linewidth]{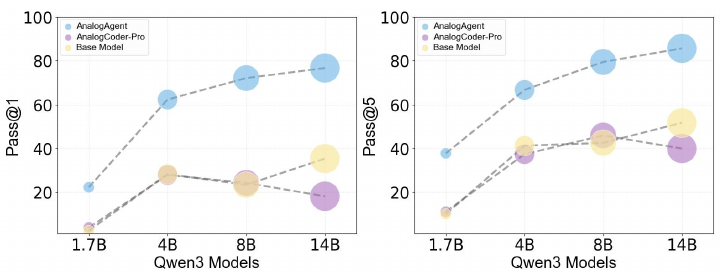}
  \caption{Performance comparison of Qwen base models, AnalogCoder-Pro, and AnalogAgent. AnalogAgent consistently enhances compact Qwen models across all model scales.}
  \label{fig:small}
  \vspace{-0.15in}
\end{figure}

Commercial IC design often cannot use APIs to access cloud-hosted LLMs because PDKs are proprietary and protected under non-disclosure agreements (NDAs). Therefore, it is important to maintain agent efficiency even when using smaller models, which are more suitable for on-premises or locally hosted servers.
To this end, we replace the backbone LLMs with compact open-weight models (i.e., \texttt{Qwen3-1.7B}, \texttt{Qwen3-4B-Instruct}, \texttt{Qwen3-8B}, and \texttt{Qwen3-14B}) while keeping the agentic workflows unchanged. 
As shown in Fig.~\ref{fig:small}, AnalogAgent achieves higher success rates and solves more tasks than both AnalogCoder-Pro and the corresponding base models on all LLM scales. At the 1.7B scale, Pass@1 improves from 2.4 to 22.2. At 4B and 8B, Pass@1 improves from 28.2$\rightarrow$62.3 and 23.3$\rightarrow$72.1, respectively. At 14B, Pass@1 further improves from 35.3$\rightarrow$76.7. In contrast, AnalogCoder-Pro does not consistently improve performance on compact LLMs and can even degrade performance in certain cases (e.g., Qwen3-14B). Overall, improvements remain consistent across model scales and become more pronounced for larger backbones, indicating that execution-driven feedback and Self-Evolving Memory amplify, rather than substitute for, backbone model capacity. Full results are reported in \textcolor{blue}Appendix \ref{appendix G}.
These findings lend empirical support to the potential of small agents as an efficiency-oriented strategy for practical deployments \cite{belcak2025small} and further underscore that the benefits of AnalogAgent are not merely a consequence of scaling the backbone model, but arise from the agentic reasoning of iterative feedback and the accumulation and retrieval of reusable design knowledge.
From a practical standpoint, this suggests a cost-effective path for industry adoption, as many organizations prefer to deploy compact open-weight models on local servers for privacy, controllability, and reduced inference cost.

\section{Conclusion and Future Work}

In this work, we propose a training-free agentic framework for analog circuit design automation that seamlessly integrates an LLM-based multi-agent system with a self-evolving memory mechanism. 
By learning an Adaptive Design Playbook that accumulates reusable design knowledge, AnalogAgent effectively mitigates context attrition and surpasses existing baselines without relying on pre-specified expert knowledge, curated circuit libraries, or external databases. 
Notably, AnalogAgent substantially improves performance for small open-weight LLMs, spanning 1.7B to 14B Qwen models, indicating that high-quality design reasoning can be achieved even with compact backbones.
Looking ahead, this framework opens several avenues for advancing LLM-driven analog design. Future work will expand the evaluation to more complex analog circuits and broader real-world constraints, and will establish new benchmarks that better reflect practical design scenarios.

\bibliographystyle{ACM-Reference-Format}
\bibliography{sample-base}




\begingroup

\setlength{\parskip}{3pt}
\appendix

\clearpage


\section{Definitions and Notations}

\begingroup
\small
\setlength{\tabcolsep}{8pt}
\renewcommand{\arraystretch}{1.12}

\captionof{table}{Definitions and Notations.}
\label{tab:notation}
\centering
\noindent\textit{Notations are organized following the execution pipeline of AnalogAgent.}

\vspace{2mm}

\noindent\makebox[\linewidth][c]{%
\begin{tabular}{l p{0.70\linewidth}}
\toprule
\textbf{Notation} & \textbf{Definition} \\
\midrule
$S$ & Circuit specification (task description and pass criteria). \\
$g_i$ & Sub-goal at iteration $i$. \\
$K$ & Max attempts per task (iteration budget). \\
$N$ & Number of benchmark tasks. \\

$M$ & Long-term memory of reusable rules and fixes. \\
$E_i$ & Retrieved memory entries at iteration $i$. \\
$\Delta_i$ & Prompt directives distilled from $E_i$. \\
$\mathcal{P}_i$ & Adaptive Design Playbook assembled at iteration $i$. \\

$F_i$ & Generated executable design script at iteration $i$. \\
$L_i$ & Execution feedback from running $F_i$. \\
$O_i$ & Aggregated execution evidence up to iteration $i$. \\

$V(\cdot)$ & Validator for constraint and functional checks. \\
$v_i$ & Validation result at iteration $i$. \\
$m_i$ & Failure message if $v_i$ is false. \\
$C(\cdot)$ & Constraint checker for structural/API/naming constraints. \\
$c_i$ & Checker outcome at iteration $i$ (pass/fail and/or violations). \\
$P(\cdot)$ & Simulation procedure (e.g., operating point, DC sweep, transient). \\
$W_i$ & Waveform evidence at iteration $i$ (if applicable). \\

$\delta_i$ & Targeted fix suggested by the Design Optimizer. \\
$r_i$ & Curated rule written back to memory. \\

$f_{\text{success}}$ & Whether the task is solved. \\
$T_{\text{TTFS}}$ & Time-to-first-success (elapsed time until first verified success). \\
\text{CSR}(k) & Cumulative success rate within the first $k$ attempts. \\

\bottomrule
\end{tabular}}
\endgroup

\section{\textsc{\MakeUppercase{Data and reproducibility}}}\label{appendixB}

This appendix first details the construction, collection, preprocessing, and standardization of the domain-specific benchmark dataset, 
and then describes the reproducible experimental configuration used in this work. 
Both the implementation code and benchmark data are released via an anonymous artifact repository to support 
independent verification, trajectory reconstruction, and full experimental reproduction.

\subsection{Domain-Specific Data Integration}\label{appendixB_data}

\begin{table*}[!tbp]
\centering
\setlength{\tabcolsep}{8pt}
\renewcommand{\arraystretch}{1.18}
\caption{Benchmark Descriptions. Difficulties are distinguished by background colors
(\textcolor{easytxt}{easy}, \textcolor{medtxt}{medium}, and \textcolor{hardtxt}{hard}).}
\label{type}
\small
\begin{tabular}{r l l @{\qquad} r l l}
\toprule
\textbf{Id} & \textbf{Type} & \textbf{Circuit Description}
& \textbf{Id} & \textbf{Type} & \textbf{Circuit Description} \\
\midrule
\easycell 1 & \easycell Amplifier & \easycell Single-stage CS amplifier with resistive load
& \hardcell 16 & \hardcell Oscillator & \hardcell RC phase-shift oscillator \\
\easycell 2 & \easycell Amplifier & \easycell Three-stage CS amplifier with resistive loads
& \hardcell 17 & \hardcell Oscillator & \hardcell Wien Bridge oscillator \\
\easycell 3 & \easycell Amplifier & \easycell Common-drain (source follower) amplifier
& \hardcell 18 & \hardcell Integrator & \hardcell Op-amp integrator \\
\easycell 4 & \easycell Amplifier & \easycell Single-stage common-gate amplifier
& \hardcell 19 & \hardcell Differentiator & \hardcell Op-amp differentiator \\
\easycell 5 & \easycell Amplifier & \easycell Single-stage NMOS cascode amplifier
& \hardcell 20 & \hardcell Adder & \hardcell Op-amp adder \\
\easycell 6 & \easycell Inverter & \easycell NMOS inverter with resistive load
& \hardcell 21 & \hardcell Subtractor & \hardcell Op-amp subtractor \\
\easycell 7 & \easycell Inverter & \easycell CMOS logical inverter
& \hardcell 22 & \hardcell Schmitt & \hardcell Non-inverting Schmitt trigger \\
\easycell 8 & \easycell CurrentMirror & \easycell NMOS constant current source
& \hardcell 23 & \hardcell VCO & \hardcell Voltage-controlled oscillator \\

\medcell 9 & \medcell Amplifier & \medcell Two-stage amplifier with Miller compensation
& \hardcell 24 & \hardcell PLL & \hardcell Phase-locked loop \\
\medcell 10 & \medcell Amplifier & \medcell CS amplifier with diode-connected PMOS load
& \hardcell 25 & \hardcell Comparator & \hardcell Op-amp comparator \\
\medcell 11 & \medcell Opamp & \medcell Differential op-amp with PMOS mirror load
& \hardcell 26 & \hardcell Filter & \hardcell Passive low-pass filter \\
\medcell 12 & \medcell CurrentMirror & \medcell Cascode current mirror
& \hardcell 27 & \hardcell Filter & \hardcell Passive high-pass filter \\
\medcell 13 & \medcell Opamp & \medcell Single-stage differential CS op-amp
& \hardcell 28 & \hardcell BandPass & \hardcell Passive band-pass filter \\

\hardcell 14 & \hardcell Opamp & \hardcell Two-stage differential op-amp with active loads
& \hardcell 29 & \hardcell BandStop & \hardcell Passive band-stop filter \\
\hardcell 15 & \hardcell Opamp & \hardcell Telescopic cascode op-amp
& \hardcell 30 & \hardcell Mixer & \hardcell Gilbert cell mixer \\
\bottomrule
\end{tabular}
\end{table*}

\noindent\textbf{Dataset Origin and Domain-Specific Motivation.}
The benchmark used in this work is constructed from domain-specific analog circuit design datasets originally introduced in prior LLM-based analog design frameworks, including AnalogCoder~\cite{lai2025analogcoder} and AnalogCoder-Pro~\cite{lai2025analogcoder_pro}. These prior works established curated benchmark suites for evaluating code-generation-based circuit synthesis under executable simulation constraints.

Analog circuit design represents a highly domain-specific scientific workflow, where usable data must capture circuit topology, device-level behavior, simulation constraints, and verification criteria simultaneously. General-purpose code corpora or synthetic programming datasets cannot provide sufficient supervision signals for executable circuit synthesis. As a result, the benchmark follows a domain-specific data design principle, emphasizing executable validity, simulator-grounded feedback, and physics-consistent verification signals.

\noindent\textbf{Data Collection and Integration.}
The final benchmark consists of 30 circuit synthesis tasks derived by consolidating and extending task instances from the AnalogCoder and AnalogCoder-Pro benchmark suites. Task definitions were collected from publicly described benchmark configurations and reconstructed into a unified task specification format. The integrated benchmark spans multiple circuit families, including amplifiers, current mirrors, op-amps, oscillators, filters, comparators, and mixed-signal building blocks.

During integration, task definitions were normalized to ensure consistent specification structure, verification criteria, and execution interfaces. When multiple source benchmarks contained semantically overlapping circuit types, representative task variants were selected based on topology diversity and simulation difficulty, ensuring coverage across open-loop, closed-loop, and multi-block circuit categories.

\noindent\textbf{Data Preprocessing and Normalization.}
All collected task specifications were preprocessed to ensure compatibility with a unified execution and evaluation pipeline. Preprocessing includes:

\begin{itemize}
\item Standardizing circuit specification format, including required node naming conventions and interface definitions;
\item Normalizing simulation configuration, including DC operating point setup, transient analysis configuration, and sweep parameter definitions;
\item Converting verification criteria into executable checker logic, enabling automated pass/fail evaluation;
\item Aligning device model references and simulation library dependencies across tasks.
\end{itemize}

These preprocessing steps ensure that all tasks can be executed under identical runtime orchestration conditions, eliminating evaluation variance introduced by heterogeneous simulation setups.

\noindent\textbf{Data Management and Artifact Organization.}
All benchmark tasks are managed using a per-task workspace design. Each workspace stores structured artifacts including task specifications, generated design scripts, simulator execution logs, verification outputs, and optional waveform visualizations. This artifact-level management enables full reconstruction of execution trajectories and supports detailed failure diagnosis analysis.
Structured logging is used to record iteration-level execution evidence, including simulator convergence behavior, checker violations, and runtime error signatures. This design ensures that both successful and unsuccessful trajectories can be analyzed systematically.

\noindent\textbf{Data Analysis and Evaluation Signals.}
Benchmark performance is analyzed using execution-grounded metrics computed across all tasks under a fixed iteration budget. Primary evaluation metrics include Pass@$k$, cumulative success rate, and Time-to-First-Success. In addition, structured execution evidence is used to analyze failure modes, convergence behavior, and repair effectiveness across iterations.
For waveform-sensitive circuits such as oscillators and transient-dependent designs, waveform outputs are additionally generated and used for functional verification and diagnostic interpretation. These signals provide domain-specific evaluation evidence beyond static code correctness.

\noindent\textbf{Extension Beyond Prior Benchmarks.}
The benchmark used in this work is constructed by consolidating domain-specific circuit synthesis tasks from AnalogCoder and AnalogCoder-Pro into a unified specification and execution framework.
The primary contribution is not the introduction of new benchmark tasks, but the standardization of execution interfaces, verification pipelines, and evaluation protocols, enabling systematic study of execution-driven iterative circuit synthesis under a multi-agent feedback-repair paradigm.
This unified configuration enables consistent evaluation of code generation quality, execution-grounded diagnosis, and iterative repair effectiveness across heterogeneous model backbones and runtime settings.

As shown in Table~\ref{type}, the benchmark consists of 30 domain-specific circuit synthesis tasks spanning multiple circuit families. Difficulty labels are assigned based on circuit topology complexity, feedback structure, and sensitivity to simulation convergence behavior.

\subsection{Implementation and Reproducible Experimental Configuration}

This section summarizes key implementation mechanisms and runtime configurations required to reproduce the AnalogAgent experimental pipeline. Detailed source code, runtime scripts, and reproduction instructions are provided in the anonymous artifact repository on github:

\begin{center}
\textcolor{blue}{\url{https://github.com/artifact-repro/analogagent-artifact}}
\end{center}

\noindent\textbf{Execution Environment and Reproducibility.}
All experiments were conducted on Ubuntu 20.04.6 LTS with Python 3.10 using a dedicated Conda environment. 
The machine learning runtime stack includes CUDA-enabled PyTorch, vLLM for high-throughput inference serving, 
and Transformers-based model integration. Experiments were executed on a dual-GPU workstation equipped with 
two NVIDIA RTX A6000 GPUs using CUDA 12.4 and NVIDIA driver 550.54.

To ensure reproducibility, all experiments were executed under fixed iteration budgets and standardized task 
specifications. Random seeds were fixed where applicable. For each task, the system records intermediate artifacts 
including generated design scripts, execution logs, verification outputs, and waveform plots when required. 
These artifacts enable full reconstruction of execution trajectories for both successful and unsuccessful runs, 
supporting detailed failure analysis and independent verification.

\noindent\textbf{Execution-Driven Multi-Agent Runtime.}
AnalogAgent is implemented as an execution-grounded multi-agent orchestration pipeline operating on executable circuit artifacts. The runtime integrates specification-conditioned code generation, structured execution feedback collection, rule-level retrieval from Self-Evolving Memory (SEM), and curator-controlled rule consolidation. Internal prompts act as structured interfaces for transporting sub-goals, execution observations, and retrieved rule-level constraints rather than as standalone knowledge sources. Circuit-domain constraints and repair heuristics are stored in SEM and injected through the Adaptive Design Playbook during runtime.

During each iteration, the runtime constructs a task-specific instruction context combining circuit specification, current sub-goal derived from execution observations, and retrieved rule-level SEM entries. Execution feedback including simulator logs, checker outputs, operating-point diagnostics, and waveform evidence is normalized into structured observations and consumed by the Design Optimizer and Knowledge Curator for targeted repair and rule consolidation.

\noindent\textbf{Multi-Backbone Support.}
AnalogAgent is backbone-agnostic and supports both local open-weight inference and API-based inference. Local experiments use vLLM-based serving, while cloud inference uses vendor SDK integrations. Evaluated backbone families include Qwen3 series (local deployment), Gemini API models, and GPT-series API models. All models are evaluated under identical task settings, iteration budgets, and execution-feedback pipelines.


\section{\textsc{\MakeUppercase{AnalogAgent Inference Workflow}}}\label{appendixC_workflow}

This appendix details the end-to-end inference workflow of AnalogAgent, describing how multi-agent execution feedback, Self-Evolving Memory (SEM), and Adaptive Design Playbook–guided generation interact during iterative circuit synthesis.

\subsection{System Overview and Core Components}

AnalogAgent is an execution-driven agentic framework for automated analog circuit design. The system couples iterative code execution, multi-agent diagnosis, and long-term experience accumulation to improve correctness, robustness, and convergence efficiency across heterogeneous circuit tasks.
At a high level, AnalogAgent operates through three tightly coupled layers:

\textbf{(1) Self-Evolving Memory (SEM).}
SEM is a persistent rule-level long-term memory that stores curated and transferable knowledge distilled from historical execution trajectories. Instead of storing raw traces or full prompts, SEM maintains compact structured rules such as validated repair patterns, topology-specific constraints, and stable checker-compliant design practices. SEM corresponds to the long-term memory $M$ used during inference.

\textbf{(2) Adaptive Design Playbook.}
The Adaptive Design Playbook is a per-iteration instruction layer constructed dynamically by retrieving relevant entries from SEM and combining them with iteration-specific design constraints. The playbook contains:
(i) \emph{Design Instruction}, which encodes task-specific requirements such as node naming, supply conventions, and functional verification constraints; and
(ii) \emph{Relevant Knowledge}, which includes transferable experience such as previously validated fixes and task-type-specific heuristics.
The playbook therefore acts as a task-conditioned operational view over SEM.

\textbf{(3) Multi-Agent System (MAS).}
AnalogAgent adopts a tight, three-agent refinement loop operating over executable artifacts and execution traces at runtime:
\begin{itemize}
\item \textbf{Code Generator Agent:} Generates executable PySpice circuit implementation and testbench code conditioned on the specification and playbook directives.
\item \textbf{Design Optimizer Agent:} Diagnoses execution failures using structured signals such as checker outputs, simulation diagnostics, and waveform evidence, and proposes concrete, targeted fix directives.
\item \textbf{Knowledge Curator Agent:} Converts evidence-backed fixes into transferable rule entries through filtering, conflict checking, and deduplication before writing them into SEM.
\end{itemize}

To control computational cost and standardize evaluation, each task is associated with a fixed iteration budget (Max Attempts), which caps the number of generate–execute–refine cycles.

\subsection{Execution-Driven Iterative Inference Loop}

The AnalogAgent inference procedure follows an iterative design – execute – diagnose – refine cycle.

\noindent\textbf{Sub-goal formation and experience retrieval.}
At the beginning of each iteration, the system aggregates execution observations (e.g., checker violations, simulation results, diagnostic messages, and waveform anomalies) together with the original circuit specification to form the next actionable sub-goal. The playbook then retrieves a small set of relevant SEM entries and distills them into concise prompt directives, forming structured delta context injected into the generation prompt.

\noindent\textbf{Executable generation and validation-driven feedback.}
Conditioned on the specification and playbook directives, the Code Generator produces an executable PySpice design script. The script is executed to produce multi-source feedback signals, including:
netlist-level structural checks,
operating-point and DC sweep diagnostics,
functional verification outcomes,
and waveform evidence when applicable.
A validation module enforces structural correctness, connectivity integrity, and task-level functional constraints. When validation fails, failure messages are propagated as structured feedback for the next iteration rather than triggering full prompt re-generation.

\noindent\textbf{Failure diagnosis, rule curation, and memory evolution.}
The Design Optimizer analyzes full execution trajectories (code, logs, checker output, and waveform evidence) to localize root causes and produce minimal targeted fix directives. The Knowledge Curator then converts evidence-backed fixes into admission-controlled rule entries and writes them into SEM. This separation between short-term iterative refinement and long-term rule consolidation prevents prompt bloat and preserves stable task-specific constraints across successive iterations.

\noindent\textbf{Workspace materialization and traceability.}
For reproducibility and debugging, the system materializes per-iteration workspace artifacts, including task specification, sub-goal history, retrieved memory entries, execution logs, waveform plots, and generated executable scripts.

A formal algorithmic description is provided in Algorithm~\ref{alg:analogagent_infer}.
\subsection{Multi-Agent Execution Process Visualization}

Figure~\ref{fig:w} provides a system-level visualization of the multi-agent execution loop. The Code Generator produces executable circuit implementations under specification and playbook guidance. The Design Optimizer closes the execution feedback loop by performing requirement validation, simulation diagnostics, and curve-level waveform analysis when applicable, and converts failure evidence into targeted fix directives. The Knowledge Curator aggregates diverse multimodal execution evidence and writes conflict-checked transferable rules into SEM, enabling cross-iteration and task-level knowledge reuse.



\FloatBarrier
\section{Algorithmic Procedure}

\begin{algorithm}
\footnotesize
\caption{AnalogAgent Inference Process}
\label{alg:analogagent_infer}

\KwIn{Circuit specification $S$}
\KwOut{Final executable design script $F_{\text{exec}}$ and verified outputs}

Initialize aggregated observations $O \leftarrow \emptyset$\;
$f_{\text{success}} \leftarrow \text{false}$\;
attempts $\leftarrow 0$\;

\While{$\neg f_{\text{success}}$ \textbf{and} attempts $< K$}{

  \tcp{\textit{Memory-guided planning}}
  Form the next actionable sub-goal $g$ from $S$ and $O$\;
  Retrieve relevant memory entries $E$ (design rules, constraints, and validated repair patterns)\;
  Distill $E$ into concise directives $\Delta$\;

  \tcp{\textit{Code generation and execution}}
  Generate executable PySpice script
  $F_{\text{exec}} \leftarrow \textsc{CodeGenerator}(S, g, \Delta)$\;
  Execute $F_{\text{exec}}$ to obtain feedback $L$
  (checker diagnostics, simulation outcomes, and verification signals)\;
  Update observations $O \leftarrow O \cup \{L\}$\;

  \tcp{\textit{Validation and diagnosis}}
  $v, m \leftarrow \textsc{Validate}(S, O)$\;
  \If{$\neg v$}{
    $\delta \leftarrow \textsc{DesignOptimizer}(S, F_{\text{exec}}, O, m)$\;
    Use $\delta$ to guide regeneration in the next iteration\;
  }

  \tcp{\textit{Curation and memory update}}
  \If{new actionable evidence is identified}{
    $r \leftarrow \textsc{Agent}(S, O, v)$\;
    Write $r$ into long-term memory and update the playbook\;
  }

  attempts $\leftarrow$ attempts $+ 1$\;

  \tcp{\textit{Stop criterion}}
  $f_{\text{success}} \leftarrow \textsc{IsSolved}(S, O)$\;
}

\Return final executable design script and verified outputs\
\end{algorithm}

The AnalogAgent inference process is summarized in Algorithm~\ref{alg:analogagent_infer}. 
Given a circuit specification, the system performs iterative memory-guided synthesis through a 
generate--execute--diagnose--refine loop under a fixed attempt budget. 

At each iteration, the system constructs an Adaptive Design Playbook by retrieving transferable 
rule-level knowledge from Self-Evolving Memory and distilling it into actionable generation 
directives. The Code Generator produces an executable circuit implementation conditioned on the 
specification and playbook constraints, and execution feedback is collected through structural 
checks, simulation diagnostics, and functional verification. When validation fails, the Design 
Optimizer proposes localized repair directives that guide regeneration in subsequent iterations. 
Evidence-backed repair patterns are optionally consolidated into memory through curator-controlled 
filtering and deduplication, enabling cross-task knowledge accumulation without rewriting existing 
rules.

The inference loop terminates when a verified design is obtained or when the maximum attempt budget 
is reached. Because memory updates are strictly incremental and retrieval is deterministic, the 
framework maintains stable constraint conditioning across iterations while allowing new 
evidence-backed knowledge to be incorporated over time.

\FloatBarrier

\begin{figure*}[t]
  \centering
  \includegraphics[width=\textwidth]{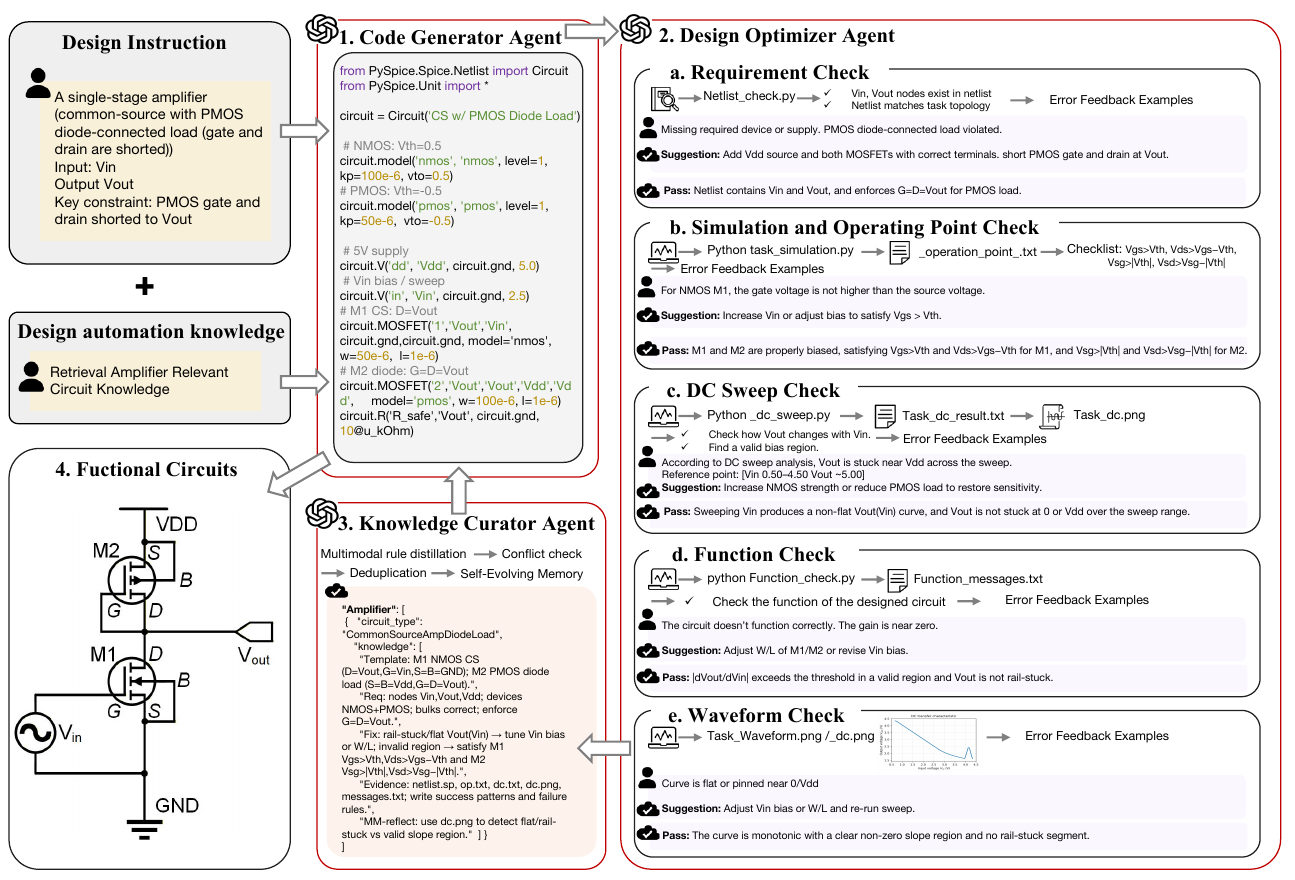}
  \caption{\textbf{AnalogAgent workflow with MAS--SEM coupling.}
The system solves each task by iterating an inner \emph{Multi-Agent System (MAS)} loop
(Code Generator $\rightarrow$ Design Optimizer $\rightarrow$ Knowledge Curator) over executable PySpice artifacts.
Execution feedback (checker logs, operating-point/DC-sweep diagnostics, and waveform evidence when available)
is used for diagnosis and targeted fixes.
The curator distills evidence-backed rules with conflict checking and deduplication,
writes transferable entries into \emph{Self-Evolving Memory (SEM)}, and the retrieved rules are injected as
playbook-style constraints to guide subsequent iterations and related tasks.}
  \label{fig:w}
\end{figure*}

\section{\textsc{\MakeUppercase{Context Attrition Case Study on Task 25)}}}
\label{CONTEXT ATTRITION CASE STUDY}
\subsection{Checkpoint Evidence of Context Attrition}
\noindent
\begin{tcolorbox}[
  enhanced,
  breakable,
  width=\columnwidth,
  colback=gray!10,
  colframe=black,
  title=Context Attrition Case Study (Task 25): Checkpoint Evidence,
  boxrule=0.5mm,
  left=2mm,
  right=2mm
]
\label{structure1}
\small
\ttfamily
\textbf{[Representative Iterations]}\\
\textbf{it1}: \textbf{4.1K} prompt tokens \\
\textbf{it10}: \textbf{2.4K} prompt tokens \\
\textbf{it30}: \textbf{1.4K} prompt tokens \\[2pt]

\textbf{[Attrition Pattern]}\\
\textbf{Type-1 (Executable feedback $\rightarrow$ removed):} simulator-level convergence traces are progressively pruned.\\
\textbf{Type-2 (Concrete diagnosis $\rightarrow$ abstract rule):} failure mode evidence is replaced by generic MOSFET operating principles.\\
\textbf{Type-3 (Circuit semantics $\rightarrow$ surface exception):} the retained context collapses to a Python runtime error without analog debug cues.\\[2pt]

\textbf{[Evidence @it1: Simulator-grounded diagnostics]}\\
-- \emph{Failure mode indicator:} ``Warning: singular matrix: check nodes vin and vin''\\
-- \emph{Solver trajectory:} gmin stepping $\rightarrow$ dynamic gmin stepping $\rightarrow$ source stepping\\
-- \emph{Termination signals:} ``doAnalyses: iteration limit reached''; ``run simulation(s) aborted''\\[2pt]

\textbf{[Evidence @it10: Abstracted operating-point rules]}\\
-- \emph{Rule-of-thumb constraints:} ``ensure $V_{GS} > V_{TH}$''; ``ensure $V_{DS} > V_{GS}-V_{TH}$''\\
-- \emph{Loss:} convergence trace and node-level singularity context no longer retained as an executable debugging trajectory\\[2pt]

\textbf{[Evidence @it30: Surface-level exception]}\\
-- \emph{Runtime symptom only:} ``KeyError: `Source\_M1\_M2\,'\,''\\
-- \emph{Loss:} no simulator-level evidence (e.g., singular matrix / stepping / abort reason) to diagnose the original circuit failure\\[2pt]

\textbf{[Conclusion: Context Attrition]}\\
Across iterations, the prompt compresses from it1 to it30 while shifting away from simulator-grounded evidence. 
As a result, key task information and diagnostic cues are \emph{no longer retained in full}, leaving an underspecified context that weakens targeted debugging and subsequent corrections.

\end{tcolorbox}
\subsection{Mechanism and Impact of Context Attrition}
In Task 25, context attrition manifests as a shift from simulator-grounded executable feedback (convergence traces) to abstract rule-of-thumb diagnostics and finally to a superficial runtime exception, progressively erasing the domain-specific evidence.

We present a checkpoint-based case study to illustrate context attrition under iterative LLM prompting. On Task 25, the retained context exhibits a pronounced brevity bias, with the prompt tokens shrinking from 4.1k at it1 to 1.4k at it30, a 65.9\% reduction relative to it1. This compression is not merely stylistic, as it progressively removes executable feedback that encodes domain-specific debugging value. At it1, the prompt preserves simulator-level evidence, including ngspice convergence traces such as gmin stepping, explicit failure indicators such as singular matrix and check nodes, escalation events such as dynamic gmin stepping failed and starting source stepping, and the termination rationale such as iteration limit reached and simulation aborted. By it10, the prompt is shorter and increasingly abstracts these trajectories into higher-level guidance, retaining task keywords such as Vin, Vref, Vout, sweep, and comparator while omitting parts of the diagnostic chain. By it30, the long-form convergence trace is no longer retained, leaving mainly high-level task framing without the fine-grained simulator feedback required for targeted fixes. These checkpoints concretely demonstrate context attrition, where repeated rewrites compress and smooth specialized diagnostic details, thereby weakening the prompt as a carrier of domain expertise across iterations.
The Case~25 trajectory also demonstrates how failure evidence can be distilled into reusable diagnostic constraints through SEM-mediated consolidation, grounded in concrete runtime observations rather than hypothetical cross-task transfer scenarios.

\section{\textsc{\MakeUppercase{Self-Evolving Memory and Adaptive Design Playbook}}}

\subsection{Memory Entry Semantics}
Self-Evolving Memory (SEM) stores \emph{transferable diagnostic rules} distilled from execution trajectories, rather than free-form narrative summaries. Each entry is intended to be actionable in subsequent runs: it links an observable \emph{failure signature} (or other salient execution evidence) to a corresponding \emph{corrective heuristic} that can be applied as a localized patch. The primary evidence sources include simulator logs, checker outputs, operating-point and sweep diagnostics, and (when applicable) waveform plots. This evidence-grounded representation encourages precise reuse: later iterations can condition generation on concrete ``what to check'' and ``what to change'' guidance under specified observed conditions, instead of re-deriving procedures from scratch.

For consistency and auditability, entries are kept atomic and scoped. Operationally, an entry can be read as a compact mapping:
\emph{Trigger} (task type and/or observed signature) $\rightarrow$ \emph{Evidence} (what was observed) $\rightarrow$ \emph{Rule/Patch} (what to enforce or modify) $\rightarrow$ \emph{Applicability} (when the rule should be invoked). Atomic scoping also supports compositional retrieval: a small subset of entries can be assembled into iteration directives without expanding the prompt with broad, loosely relevant text. Importantly, SEM primarily accumulates failure-derived corrective rules rather than full successful circuit designs. Successful designs are used as short-term guidance within the same task but are not directly written into SEM as reusable rules unless they expose transferable structural constraints or stability guarantees. This design choice prevents overfitting memory to task-specific topology details and improves cross-task generalization.

\subsection{Memory Format and Example}

\noindent
\begin{tcolorbox}[
  enhanced,
  breakable,
  width=\columnwidth,
  colback=gray!10,
  colframe=black,
  title=Curator-Processed Long-Term Memory Structure Illustration,
  boxrule=0.5mm,
  left=2mm,
  right=2mm
]
\label{structure2}
\small
\ttfamily
\textbf{[General Rules]}\\
- Outputs must be named exactly: Voutp and Vout.\\
- NMOS bulk must tie to source; PMOS bulk must tie to Vdd.\\
- CRITICAL API: Never use \texttt{circuit.add\_nodes()}; nodes are created implicitly.\\
- CRITICAL API: Capacitor initial condition must use \texttt{ic=...} (not \texttt{initial\_condition=...}).\\
- Opamp subcircuit interface: exactly three pins (non-inv, inv, out); do not redefine Opamp.\\
\\
\textbf{[Task-Type Rules: Comparator]}\\
- Subcircuit pin names must be passed as a single iterable to \texttt{subcircuit()}.\\
- Any \texttt{circuit.X(...)} instantiation requires the subcircuit to be defined/loaded beforehand.\\
\\
\textbf{[Task-Type Rules: Oscillator]}\\
- Place each Python statement on its own line (avoid concatenation causing \texttt{SyntaxError}).\\
- Apply initial conditions to active loop nodes (kickstart), not fixed supply/bias nodes.\\
- Ensure every node has a DC path to a reference for operating-point convergence.\\
\\
\textbf{[Task-Type Rules: Integrator / Differentiator]}\\
- Strictly follow explicit checker constraints on allowed op-amp modeling; prefer task-specific
requirements over generic API rules.\\
- Keep statement separation strict to avoid \texttt{SyntaxError}.\\
\end{tcolorbox}
The box illustrates the curator-processed memory structure used for retrieval, organized into general rules and task-type rules. General rules encode API constraints and invariant circuit conventions, while task-type rules capture failure signatures and corrective constraints commonly observed for a circuit family.
Not all execution experiences are written into SEM. The Knowledge Curator applies admission control before memory insertion. A candidate rule must satisfy at least one of the following criteria:
(i) resolves a repeated failure pattern across iterations,
(ii) enforces checker or simulator constraints violated in multiple attempts,
(iii) represents a stable design practice independent of specific parameter values.
This admission process prevents noisy or task-specific artifacts from polluting long-term memory.

To control memory growth, SEM stores compact rule-level entries rather than full trajectories, and lightweight deduplication and conflict filtering are applied during curation. Retrieval is bounded to a small number of high-relevance entries per iteration.

\subsection{Success vs.\ Failure Evidence Handling}
The workflow produces two categories of reusable information: (i) \emph{evidence-backed repair rules} derived from failures and verification violations, and (ii) \emph{stabilizing patterns} observed in verified runs that help maintain invariants across subsequent iterations. The update policy is intentionally conservative about what is promoted into SEM: only entries that can be stated as transferable constraints or diagnostic procedures, and that are supported by concrete execution evidence, are admitted. In contrast, task-specific incidental details that do not generalize beyond a single run are treated as short-term context (e.g., reference artifacts within the current iteration loop) rather than long-term rules.

This separation supports two objectives. First, failure-driven rules reduce repeated mistakes by encoding specific signatures and fixes (e.g., ``operating-point does not converge due to missing DC path'' $\rightarrow$ ``add a bias resistor or reference path''). Second, stabilizing patterns from verified executions preserve conventions that are easy to regress under iterative prompting (e.g., required node naming and API usage invariants). In both cases, the decision to store information is guided by transferability and evidence: the system prefers compact, auditable entries that can be retrieved and assembled into iteration directives without inflating the prompt with full code histories.

\subsection{Curator Filtering and Update Policy}
The Knowledge Curator enforces a lightweight but explicit admission pipeline before writing to SEM. Candidate entries are first checked for conflicts with invariant conventions (e.g., required node naming and API constraints) and for logical consistency with existing rules. Entries that duplicate recently stored guidance are filtered through semantic de-duplication to avoid accumulating redundant phrasing of the same constraint. Finally, accepted entries are normalized into an itemized representation so that retrieval can remain targeted and compositional.

\begin{table*}[t]
\centering
\caption{Per-task Pass@k performance comparison across small-scale LLM backbones. Results report AnalogCoder, AnalogCoder-Pro, and AnalogAgent performance on 30 benchmark circuit design tasks.}
\label{tab:small_model}
\scriptsize
\setlength{\tabcolsep}{1.6pt}
\renewcommand{\arraystretch}{1.6}
\begin{tabular}{ll cccccccccccccccccccccccccccccc}
\toprule
\textbf{Model (Avg | Solved)} & \textbf{Metric} & \textbf{1} & \textbf{2} & \textbf{3} & \textbf{4} & \textbf{5} & \textbf{6} & \textbf{7} & \textbf{8} & \textbf{9} & \textbf{10} & \textbf{11} & \textbf{12} & \textbf{13} & \textbf{14} & \textbf{15} & \textbf{16} & \textbf{17} & \textbf{18} & \textbf{19} & \textbf{20} & \textbf{21} & \textbf{22} & \textbf{23} & \textbf{24} & \textbf{25} & \textbf{26} & \textbf{27} & \textbf{28} & \textbf{29} & \textbf{30} \\
\midrule
\multirow{2}{*}{\makecell[l]{\textbf{Qwen3-1.7B} \\ \textcolor{blue!70!black}{\tiny Avg: 2.4/10.0 | Sol: 10/30}}} & Pass@1 & 0 & 0 & 0 & 3.3 & 3.3 & 3.3 & 3.3 & 0 & 0 & 13.3 & 13.3 & 3.3 & 3.3 & 23.3 & 3.3 & 0 & 0 & 0 & 0 & 0 & 0 & 0 & 0 & 0 & 0 & 0 & 0 & 0 & 0 & 0 \\
 & Pass@5 & 0 & 0 & 0 & 16.7 & 16.7 & 16.7 & 16.7 & 0 & 0 & 53.8 & 53.8 & 16.7 & 16.7 & 76.4 & 16.7 & 0 & 0 & 0 & 0 & 0 & 0 & 0 & 0 & 0 & 0 & 0 & 0 & 0 & 0 & 0 \\
\hline
\multirow{2}{*}{\makecell[l]{\textbf{AnalogCoder-Pro (Qwen3-1.7B)} \\ \textcolor{blue!70!black}{\tiny Avg: 3.8/11.0 | Sol: 7/30}}} & Pass@1 & 0 & 0 & 0 & 3.3 & 10 & 0 & 0 & 3.3 & 0 & 0 & 26.7 & 10 & 6.7 & 53.3 & 0 & 0 & 0 & 0 & 0 & 0 & 0 & 0 & 0 & 0 & 0 & 0 & 0 & 0 & 0 & 0 \\
 & Pass@5 & 0 & 0 & 0 & 16.7 & 43.3 & 0 & 0 & 16.7 & 0 & 0 & 81.5 & 43.3 & 31 & 98.6 & 0 & 0 & 0 & 0 & 0 & 0 & 0 & 0 & 0 & 0 & 0 & 0 & 0 & 0 & 0 & 0 \\
\hline
\multirow{2}{*}{\makecell[l]{\textbf{AnalogAgent (Qwen3-1.7B)} \\ \textcolor{blue!70!black}{\tiny Avg: 22.2/37.7 | Sol: 13/30}}} & Pass@1 & 53.3 & 0 & 0 & 56.7 & 33.3 & 80 & 3.3 & 33.3 & 0 & 90 & 43.3 & 83.3 & 10 & 76.7 & 80 & 0 & 0 & 0 & 0 & 0 & 0 & 0 & 0 & 0 & 0 & 53.3 & 0 & 0 & 0 & 0 \\
 & Pass@5 & 98.6 & 0 & 0 & 99.1 & 89.1 & 99.9 & 16.7 & 89.1 & 0 & 100 & 95.5 & 99.9 & 43.3 & 99.9 & 99.9 & 0 & 0 & 0 & 0 & 0 & 0 & 0 & 0 & 0 & 0 & 98.6 & 0 & 0 & 0 & 0 \\
\hline
\multirow{2}{*}{\makecell[l]{\textbf{Qwen3-4B} \\ \textcolor{blue!70!black}{\tiny Avg: 28.2/41.2 | Sol: 13/30}}} & Pass@1 & 23.3 & 0 & 0 & 56.7 & 0 & 66.7 & 86.7 & 100 & 30 & 100 & 93.3 & 66.7 & 40 & 70 & 0 & 0 & 0 & 0 & 0 & 0 & 0 & 0 & 0 & 0 & 0 & 0 & 0 & 0 & 86.7 & 26.7 \\
 & Pass@5 & 76.4 & 0 & 0 & 99.1 & 0 & 99.8 & 100 & 100 & 85.7 & 100 & 100 & 99.8 & 93.9 & 99.9 & 0 & 0 & 0 & 0 & 0 & 0 & 0 & 0 & 0 & 0 & 0 & 0 & 0 & 0 & 100 & 81.5 \\
\hline
\multirow{2}{*}{\makecell[l]{\textbf{AnalogCoder-Pro (Qwen3-4B)} \\ \textcolor{blue!70!black}{\tiny Avg: 27.8/37.3 | Sol: 14/30}}} & Pass@1 & 70 & 63.3 & 0 & 73.3 & 3.3 & 100 & 83.3 & 93.3 & 0 & 3.3 & 83.3 & 86.7 & 50 & 100 & 20 & 0 & 0 & 0 & 3.3 & 0 & 0 & 0 & 0 & 0 & 0 & 0 & 0 & 0 & 0 & 0 \\
 & Pass@5 & 99.9 & 99.7 & 0 & 100 & 16.7 & 100 & 100 & 100 & 0 & 16.7 & 100 & 100 & 97.9 & 100 & 70.2 & 0 & 0 & 0 & 16.7 & 0 & 0 & 0 & 0 & 0 & 0 & 0 & 0 & 0 & 0 & 0 \\
\hline
\multirow{2}{*}{\makecell[l]{\textbf{AnalogAgent (Qwen3-4B)} \\ \textcolor{blue!70!black}{\tiny Avg: 62.3/66.7 | Sol: 20/30}}} & Pass@1 & 100 & 100 & 100 & 100 & 100 & 100 & 100 & 100 & 86.7 & 100 & 100 & 100 & 100 & 93.3 & 83.3 & 0 & 0 & 0 & 66.7 & 0 & 0 & 0 & 0 & 0 & 0 & 93.3 & 70 & 0 & 76.7 & 100 \\
 & Pass@5 & 100 & 100 & 100 & 100 & 100 & 100 & 100 & 100 & 100 & 100 & 100 & 100 & 100 & 100 & 99.9 & 0 & 0 & 0 & 99.8 & 0 & 0 & 0 & 0 & 0 & 0 & 100 & 99.9 & 0 & 99.9 & 100 \\
\hline
\multirow{2}{*}{\makecell[l]{\textbf{Qwen3-8B} \\ \textcolor{blue!70!black}{\tiny Avg: 23.3/42.5 | Sol: 15/30}}} & Pass@1 & 40 & 3.3 & 26.7 & 60 & 26.7 & 43.3 & 56.7 & 96.7 & 3.3 & 53.3 & 50 & 46.7 & 0 & 70 & 63.3 & 0 & 0 & 60 & 0 & 0 & 0 & 0 & 0 & 0 & 0 & 0 & 0 & 0 & 0 & 0 \\
 & Pass@5 & 93.9 & 16.7 & 81.2 & 99.5 & 81.5 & 95.5 & 99.1 & 100 & 16.7 & 98.6 & 97.3 & 96.6 & 0 & 99.9 & 99.8 & 0 & 0 & 99.5 & 0 & 0 & 0 & 0 & 0 & 0 & 0 & 0 & 0 & 0 & 0 & 0 \\
\hline
\multirow{2}{*}{\makecell[l]{\textbf{AnalogCoder-Pro (Qwen3-8B)} \\ \textcolor{blue!70!black}{\tiny Avg: 24.2/45.9 | Sol: 16/30}}} & Pass@1 & 33.3 & 43.3 & 36.7 & 90 & 33.3 & 40 & 40 & 96.7 & 0 & 6.7 & 40 & 56.7 & 50 & 66.7 & 60 & 0 & 0 & 30 & 3.3 & 0 & 0 & 0 & 0 & 0 & 0 & 0 & 0 & 0 & 0 & 0 \\
 & Pass@5 & 89.1 & 95.7 & 91.8 & 100 & 89.1 & 94 & 94 & 100 & 0 & 31 & 94 & 99.1 & 97.9 & 99.8 & 99.4 & 0 & 0 & 85.7 & 16.7 & 0 & 0 & 0 & 0 & 0 & 0 & 0 & 0 & 0 & 0 & 0 \\
\hline
\multirow{2}{*}{\makecell[l]{\textbf{AnalogAgent (Qwen3-8B)} \\ \textcolor{blue!70!black}{\tiny Avg: 72.1/79.5 | Sol: 24/30}}} & Pass@1 & 100 & 100 & 100 & 100 & 93.3 & 100 & 96.7 & 100 & 96.7 & 100 & 90 & 100 & 100 & 80 & 96.7 & 0 & 76.7 & 63.3 & 100 & 0 & 0 & 50 & 33.3 & 0 & 0 & 100 & 93.3 & 0 & 93.3 & 100 \\
 & Pass@5 & 100 & 100 & 100 & 100 & 100 & 100 & 100 & 100 & 100 & 100 & 100 & 100 & 100 & 99.9 & 100 & 0 & 99.9 & 99.8 & 100 & 0 & 0 & 97.3 & 89.1 & 0 & 0 & 100 & 100 & 0 & 100 & 100 \\
\hline
\multirow{2}{*}{\makecell[l]{\textbf{Qwen3-14B} \\ \textcolor{blue!70!black}{\tiny Avg: 35.3/51.7 | Sol: 18/30}}} & Pass@1 & 90.0 & 3.3 & 96.7 & 83.3 & 93.3 & 83.3 & 96.7 & 56.7 & 50.0 & 76.7 & 63.3 & 70.0 & 63.3 & 40.0 & 90.0 & 0 & 0 & 10.0 & 0 & 0 & 0 & 0 & 0 & 0 & 0 & 0 & 0 & 0 & 20.0 & 6.7 \\
 & Pass@5 & 100 & 16.7 & 100 & 100 & 100 & 100 & 100 & 99.1 & 97.9 & 100 & 99.7 & 99.9 & 99.7 & 94.0 & 100 & 0 & 0 & 43.3 & 0 & 0 & 0 & 0 & 0 & 0 & 0 & 0 & 0 & 0 & 70.2 & 31.0 \\
\hline
\multirow{2}{*}{\makecell[l]{\textbf{AnalogCoder-Pro (Qwen3-14B)} \\ \textcolor{blue!70!black}{\tiny Avg: 18.1/39.9 | Sol: 18/30}}} & Pass@1 & 73.3 & 53.3 & 36.7 & 83.3 & 30 & 50 & 50 & 10 & 3.3 & 3.3 & 26.7 & 33.3 & 23.3 & 13.3 & 20 & 0 & 0 & 23.3 & 3.3 & 0 & 6.7 & 0 & 0 & 0 & 0 & 0 & 0 & 0 & 0 & 0 \\
 & Pass@5 & 100 & 98.6 & 91.8 & 100 & 85.7 & 97.9 & 97.9 & 43.3 & 16.7 & 16.7 & 81.5 & 89.1 & 76.4 & 53.8 & 70.2 & 0 & 0 & 76.4 & 16.7 & 0 & 31 & 0 & 0 & 0 & 0 & 0 & 0 & 0 & 0 & 0 \\
\hline
\multirow{2}{*}{\makecell[l]{\textbf{AnalogAgent (Qwen3-14B)} \\ \textcolor{blue!70!black}{\tiny Avg: 76.7/85.7 | Sol: 26/30}}} & Pass@1 & 100 & 100 & 100 & 100 & 100 & 93.3 & 100 & 100 & 96.7 & 80.0 & 73.3 & 100 & 86.7 & 80.0 & 100 & 66.7 & 0 & 100 & 70.0 & 66.7 & 33.3 & 100 & 83.3 & 0 & 0 & 100 & 73.3 & 0 & 96.7 & 100 \\
 & Pass@5 & 100 & 100 & 100 & 100 & 100 & 100 & 100 & 100 & 100 & 100 & 100 & 100 & 100 & 100 & 100 & 99.8 & 0 & 100 & 99.9 & 99.8 & 89.1 & 100 & 100 & 0 & 0 & 100 & 100 & 0 & 100 & 100 \\
\hline
\bottomrule
\end{tabular}
\end{table*}

Updates are incremental: newly admitted entries are appended as additional rules rather than triggering wholesale rewriting of the memory. This conservative update policy is designed to preserve stable, previously validated conventions while still incorporating new, evidence-backed heuristics as they arise during execution. As a result, the playbook built from retrieved entries remains interpretable and reproducible across iterations: the same core constraints persist, and only localized, verifiable additions are introduced over time.

\section{\textsc{\MakeUppercase{Additional Results on Compact LLMs}}}
\label{appendix G}
Table~\ref{tab:small_model} reports detailed per-task results under small and mid-scale backbone models. This table is provided to complement the main experimental results and to demonstrate that the performance gains of AnalogAgent are not restricted to large-capacity models, but remain consistent across a wide range of backbone sizes. All models are evaluated under the same iterative execution-driven synthesis pipeline and identical attempt budgets. Pass@$k$ is computed over iterative attempts rather than independent sampling, and therefore reflects the cumulative probability that a task is solved within the first $k$ execution-refinement iterations. For each model, ``Avg'' denotes the average Pass@1 across all tasks, while ``Solved'' indicates the number of benchmark tasks for which at least one valid solution is found within the maximum attempt budget. Columns 1--30 correspond to individual benchmark tasks. Rows group results by backbone model and framework variant, including the base model, AnalogCoder-Pro, and AnalogAgent.

\end{document}